\documentclass[a4paper, 11pt]{article}
\pdfoutput=1
\usepackage{etoolbox,verbatim}

\newtoggle{arxiv}
\toggletrue{arxiv}

\newtoggle{customthms}
\toggletrue{customthms}

\newtoggle{colt}
\togglefalse{colt}

\PassOptionsToPackage{table}{xcolor}

\usepackage{xcolor}
\usepackage{tcolorbox}
\tcbuselibrary{skins,breakable}

\tcbset{
  aibox/.style={
      width=\linewidth,
      top=6pt,
      bottom=0pt,
      left=1.5pt,
      right=1.5pt,
      colframe=black,
      colbacktitle=black,
      enhanced,
      center,
      attach boxed title to top left={yshift=-0.1in,xshift=0.15in},
      boxed title style={boxrule=0pt,colframe=white,},
    }
}
\newtcolorbox{AIbox}[2][]{aibox,title=#2,#1}

\definecolor{primalcolor}{HTML}{960000}
\definecolor{contrarycolor}{HTML}{009696}
\definecolor{primaldarkcolor}{HTML}{660000}
\definecolor{contrarydarkcolor}{HTML}{006666}
\definecolor{irongray}{HTML}{6D6E71}
\definecolor{steelgray}{HTML}{F1F4F7}
\newcommand{\linkcolor}{contrarydarkcolor}
\newcommand{\urlcolor}{contrarydarkcolor}
\newcommand{\citecolor}{contrarydarkcolor}

\newcommand{\thmcolordark}{primaldarkcolor}

\usepackage[T1]{fontenc}

\usepackage{natbib}
\usepackage{amssymb,upgreek,bm}
\usepackage{mathtools}
\usepackage[english]{babel}
\usepackage{multirow,tcolorbox,tikz-cd,turnstile,xspace,xparse}
\usepackage{relsize,appendix}
\IfFileExists{breakcites.sty}{\usepackage{breakcites}}{}
\usepackage{longtable,tablefootnote,booktabs,float,makecell}
\usepackage[normalem]{ulem}
\usepackage{tabu}
\usepackage[shortlabels]{enumitem}
\usepackage{footnote}
\usepackage{boxedminipage}
\usepackage{multirow,nicefrac}
\usepackage{verbatim} % gives the comment environment
\usepackage{nicematrix}
\usepackage{subcaption}

\usepackage[colorlinks=true,linkcolor=\linkcolor,urlcolor=\urlcolor,citecolor=\citecolor,breaklinks]{hyperref}

\usepackage{prettyref}

\usepackage[capitalise,nameinlink]{cleveref}

\iftoggle{colt}{
    \DeclareRobustCommand{\qed}{
        \usepackage{thmtools}
        \ifmmode \mathqed
        \else
            \leavevmode\unskip\penalty9999 \hbox{}\nobreak\hfill
            \quad\hbox{\qedsymbol}%
        \fi
    }
}
{
    \usepackage{amsthm}

}
\usepackage{thm-restate}

\iftoggle{arxiv}
{
    \usepackage{mathrsfs}
    \usepackage[
        a4paper,
        top=1in,
        bottom=1in,
        left=1in,
        right=1in]{geometry}
    \IfFileExists{mathdesign.sty}{%
        \usepackage[bitstream-charter]{mathdesign}%
    }{%
        \usepackage{charter}% local-preview fallback
    }
    
    \IfFileExists{PTSans.sty}{%
        \usepackage[scaled=0.92]{PTSans}%
    }{%
        \usepackage[scaled=0.92]{helvet}% local-preview fallback
    }

    \usepackage{fullpage}
    \usepackage[noend]{algpseudocode}
    \usepackage{algorithm}

}
{
    \usepackage{mathrsfs}
}

\DeclareMathAlphabet{\mathbfsf}{\encodingdefault}{\sfdefault}{bx}{n}

\numberwithin{equation}{section}

\iftoggle{arxiv}
{
    }
{
    
}
\iftoggle{arxiv}
{
    
}
{
    
}

\usepackage{thmtools}

\Crefname{equation}{Eq.}{Eqs.}
\Crefname{assumption}{Assumption}{Assumptions}
\Crefname{condition}{Condition}{Conditions}
\Crefname{claim}{Claim}{Claims}
\Crefname{property}{Property}{Properties}
\Crefname{construction}{Construction}{Constructions}

\declaretheoremstyle[
    headformat=\normalfont\textcolor{\thmcolordark}{\bfseries\NAME\,\NUMBER}\NOTE,%
    notefont={\normalfont\textcolor{\thmcolordark}{\bfseries}}, 
    notebraces={}{},
    bodyfont=\normalfont\itshape,
    spaceabove = 6pt,
    spacebelow = 6pt,
    ]{coloredthmversion}

\declaretheoremstyle[
    headformat=\normalfont\textcolor{\thmcolordark}{\bfseries\NAME\,\NUMBER}\NOTE,%
    bodyfont=\normalfont\itshape,
    spaceabove = 6pt,
    spacebelow = 6pt,
    ]{coloredthm}

\declaretheoremstyle[
    headformat=\normalfont\textcolor{\thmcolordark}{\bfseries\NAME\,\NUMBER}\NOTE,%
    bodyfont=\normalfont,
    spaceabove = 6pt,
    spacebelow = 6pt,
    ]{coloreddef}

\iftoggle{customthms}
{
    \theoremstyle{coloredthmversion}
}
{}

\iftoggle{customthms}{
  \theoremstyle{coloredthm}
  \newtheorem{theorem}{Theorem}
  \newtheorem{lemma}{Lemma}[section]
  \newtheorem{corollary}{Corollary}[section]
  \newtheorem{proposition}[lemma]{Proposition}
}
{}

\newtheorem*{thminformal*}{Informal Theorem}

\iftoggle{customthms}{
    \theoremstyle{coloreddef}
    \newtheorem{definition}{Definition}[section]

    \newtheorem{property}{Property}[section]
}
{
  
}

\newtheorem{assumption}{Assumption}[section]
\newtheorem{condition}{Condition}[section]

\makeatletter
\newcommand{\neutralize}[1]{\expandafter\let\csname c@#1\endcsname\count@}
\makeatother

\iftoggle{customthms}{
    \newtheoremstyle{named}{}{}{\itshape}{}{\bfseries}{}{.5em}{\Cref{#3} {\normalfont (informal)} }{}
    \theoremstyle{named}
    
    \theoremstyle{plain}
}
{}

\newtheorem*{theorem*}{Theorem}
\newtheorem*{lemma*}{Lemma}
\newtheorem*{corollary*}{Corollary}
\newtheorem*{proposition*}{Proposition}
\newtheorem*{claim*}{Claim}
\newtheorem*{fact*}{Fact}
\newtheorem*{observation*}{Observation}
\newtheorem*{definition*}{Definition}
\newtheorem*{remark*}{Remark}
\newtheorem*{example*}{Example}

\def\ddefloop#1{\ifx\ddefloop#1\else\ddef{#1}\expandafter\ddefloop\fi}
\def\ddef#1{\expandafter\def\csname bb#1\endcsname{\ensuremath{\mathbb{#1}}}}
\ddefloop ABCDEFGHIJKLMNOPQRSTUVWXYZ\ddefloop

\def\ddefloop#1{\ifx\ddefloop#1\else\ddef{#1}\expandafter\ddefloop\fi}
\def\ddef#1{\expandafter\def\csname frak#1\endcsname{\ensuremath{\mathfrak{#1}}}}
\ddefloop ABCDEFGHIJKLMNOPQRSTUVWXYZ\ddefloop

\def\ddefloop#1{\ifx\ddefloop#1\else\ddef{#1}\expandafter\ddefloop\fi}
\def\ddef#1{\expandafter\def\csname fr#1\endcsname{\ensuremath{\mathfrak{#1}}}}
\ddefloop ABCDEFGHIJKLMNOPQRSTUVWXYZ\ddefloop

\def\ddefloop#1{\ifx\ddefloop#1\else\ddef{#1}\expandafter\ddefloop\fi}
\def\ddef#1{\expandafter\def\csname eul#1\endcsname{\ensuremath{\EuScript{#1}}}}
\ddefloop ABCDEFGHIJKLMNOPQRSTUVWXYZ\ddefloop

\def\ddefloop#1{\ifx\ddefloop#1\else\ddef{#1}\expandafter\ddefloop\fi}
\def\ddef#1{\expandafter\def\csname scr#1\endcsname{\ensuremath{\mathscr{#1}}}}
\ddefloop ABCDEFGHIJKLMNOPQRSTUVWXYZ\ddefloop

\def\ddefloop#1{\ifx\ddefloop#1\else\ddef{#1}\expandafter\ddefloop\fi}
\def\ddef#1{\expandafter\def\csname b#1\endcsname{\ensuremath{\mathbf{#1}}}}
\ddefloop ABCDEFGHIJKLMNOPQRSTUVWXYZ\ddefloop

\def\ddefloop#1{\ifx\ddefloop#1\else\ddef{#1}\expandafter\ddefloop\fi}
\def\ddef#1{\expandafter\def\csname bhat#1\endcsname{\ensuremath{\hat{\mathbf{#1}}}}}
\ddefloop ABCDEFGHIJKLMNOPQRSTUVWXYZ\ddefloop

\def\ddefloop#1{\ifx\ddefloop#1\else\ddef{#1}\expandafter\ddefloop\fi}
\def\ddef#1{\expandafter\def\csname btil#1\endcsname{\ensuremath{\tilde{\mathbf{#1}}}}}
\ddefloop ABCDEFGHIJKLMNOPQRSTUVWXYZ\ddefloop

\def\ddefloop#1{\ifx\ddefloop#1\else\ddef{#1}\expandafter\ddefloop\fi}
\def\ddef#1{\expandafter\def\csname bst#1\endcsname{\ensuremath{\mathbf{#1}^\star}}}
\ddefloop ABCDEFGHIJKLMNOPQRSTUVWXYZ\ddefloop

\def\ddefloop#1{\ifx\ddefloop#1\else\ddef{#1}\expandafter\ddefloop\fi}
\def\ddef#1{\expandafter\def\csname bst#1\endcsname{\ensuremath{\mathbf{#1}^\star}}}
\ddefloop abcdeghijklmnopqrstuvwxyz\ddefloop

\def\ddefloop#1{\ifx\ddefloop#1\else\ddef{#1}\expandafter\ddefloop\fi}
\def\ddef#1{\expandafter\def\csname bhat#1\endcsname{\ensuremath{\hat{\mathbf{#1}}}}}
\ddefloop abcdefghijklmnopqrstuvwxyz\ddefloop

\def\ddefloop#1{\ifx\ddefloop#1\else\ddef{#1}\expandafter\ddefloop\fi}
\def\ddef#1{\expandafter\def\csname b#1\endcsname{\ensuremath{\mathbf{#1}}}}
\ddefloop abcdeghijklnopqrstuvwxyz\ddefloop

\def\ddefloop#1{\ifx\ddefloop#1\else\ddef{#1}\expandafter\ddefloop\fi}
\def\ddef#1{\expandafter\def\csname barb#1\endcsname{\ensuremath{\bar{\mathbf{#1}}}}}
\ddefloop abcdefghijklmnopqrstuvwxyz\ddefloop

\def\ddef#1{\expandafter\def\csname c#1\endcsname{\ensuremath{\mathcal{#1}}}}
\ddefloop ABCDEFGHIJKLMNOPQRSTUVWXYZ\ddefloop
\def\ddef#1{\expandafter\def\csname h#1\endcsname{\ensuremath{\widehat{#1}}}}
\ddefloop ABCDEFGHIJKLMNOPQRSTUVWXYZ\ddefloop
\def\ddef#1{\expandafter\def\csname hc#1\endcsname{\ensuremath{\widehat{\mathcal{#1}}}}}
\ddefloop ABCDEFGHIJKLMNOPQRSTUVWXYZ\ddefloop
\def\ddef#1{\expandafter\def\csname t#1\endcsname{\ensuremath{\widetilde{#1}}}}
\ddefloop ABCDEFGHIJKLMNOPQRSTUVWXYZ\ddefloop
\def\ddef#1{\expandafter\def\csname tc#1\endcsname{\ensuremath{\widetilde{\mathcal{#1}}}}}
\ddefloop ABCDEFGHIJKLMNOPQRSTUVWXYZ\ddefloop
\newcommand{\ballkr}[1][r]{\cB_{k}(r)}

\DeclareMathSymbol{\shortminus}{\mathbin}{AMSa}{"39}

\makeatletter
\newcommand\addtometadatalist[5][]{%
    \begingroup
    \if\relax#3\relax\def\sep{}\else\def\sep{#5}\fi
    \let\protect\@unexpandable@protect
    \xdef#3{\expandafter{#3}\sep #4[#1]{#2}}%
    \endgroup
}

\newcommand\metadatalist{}
\newcommand\metadataformat[2][]{{\small \textbf{#1:} #2}}
\newcommand\metadata[2][]{\addtometadatalist[#1]{#2}{\metadatalist}{\metadataformat}{\\}}
\makeatother

\newcommand{\paperwebsite}[1]{\metadata[Website]{\url{#1}}}
\newcommand{\papercode}[1]{\metadata[Code]{\url{#1}}}
\newcommand{\paperdocs}[1]{\metadata[Documentation]{\url{#1}}}
\newcommand{\paperblog}[1]{\metadata[Blog]{\url{#1}}}
\renewcommand\date[1]{\metadata[Date]{#1}}
\newcommand\correspondence[1]{\metadata[Correspondence]{#1}}
\newcommand{\taskname}[1]{{\texttt{#1}}}

\usepackage{preamble/color-edits}
\addauthor{ms}{magenta}
\addauthor{cp}{blue}

\newcommand{\ignore}[1]{}

\iftoggle{arxiv}
{
\input{preamble/arxiv_title}
}
{}

\usepackage{graphicx}   % \includegraphics (harmless if already pulled in)
\usepackage{grffile}    % allow graphics file names with spaces
\IfFileExists{bbm.sty}{%
  \usepackage{bbm}% \mathbbm{1} indicator
}{}
\providecommand{\mathbbm}[1]{\mathbb{#1}}% local-preview fallback
\usepackage{placeins}   % \FloatBarrier to keep floats in order
\usepackage{wrapfig}    % wrap text around a single-column embedded figure

\providecommand{\texorpdfstring}[2]{#1} % fallback if hyperref isn't loaded
\newcommand{\method}{\texorpdfstring{\textbf{WANDA}}{WANDA}\xspace}

\definecolor{wandahighlight}{HTML}{F3DDE0}
\newcommand{\hl}[1]{\cellcolor{wandahighlight}\textbf{#1}}

\newcommand{\guanya}[1]{}

\title{Worlds in One Demo: A Synthetic Data Engine for Learning Open-World Mobile Manipulation}

\author{\small
  Lingxiao Guo\textsuperscript{*} \quad Huanyu Li\textsuperscript{*} \quad Guanya Shi \\
  \vspace{-.3em}
  \rule{.38\textwidth}{.7pt}
  \\
  \footnotesize Carnegie Mellon University \\
  \footnotesize \textsuperscript{*}Equal contribution; order decided by a coin flip.
}

\correspondence{\href{mailto:lingxiag@andrew.cmu.edu,huanyuli@andrew.cmu.edu,guanyas@andrew.cmu.edu}{\{lingxiag,\,huanyuli,\,guanyas\}@andrew.cmu.edu}}
\paperwebsite{https://wanda.lecar-lab.org} % TODO: replace with the real project page URL

\begin{document}

\begin{tcolorbox}[
    colback=blue!60!gray!5, colframe=gray!50,
    boxrule=0pt,
    arc=2mm,
    left=1mm, right=1mm, boxsep=1mm
  ]
  \maketitle
  \vspace{-2.5em}
  \tcbline
  % Teaser shown inside the title block (under the author list)
  \begin{minipage}{\linewidth}
    \centering
    \includegraphics[width=\linewidth]{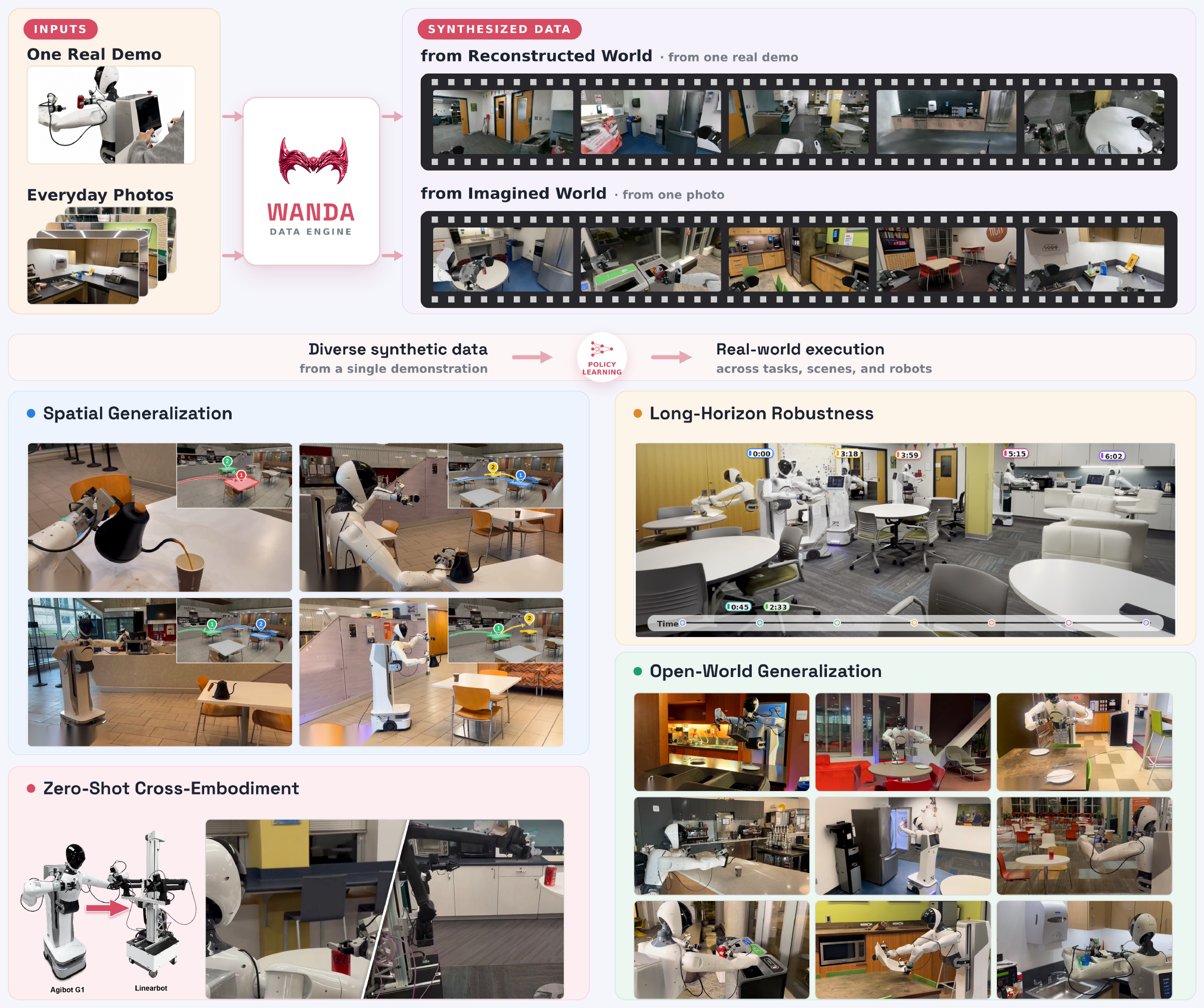}
    \captionof{figure}{%
      \method enables mobile manipulation policies achieve extreme spatial generalization,
      long horizon robustness, cross-environment generalization and cross-embodiment generalization from
      \textbf{only one human demonstration}. This is achieved by the proposed data
      engine that synthesizes diverse trajectories across reconstructed worlds and generated worlds  on different robot embodiments.%
    }
    \label{fig:teaser}
  \end{minipage}
  \tcbline
  % Metadata section (after teaser, inside tcolorbox)
  \makeatletter
  \ifdefempty{\metadatalist}{}{\metadatalist\par}
  \makeatother
\end{tcolorbox}

\begin{abstract}
Learning open-world mobile manipulation policies requires vast data to achieve spatial generalization, long-horizon robustness, and scene generalization. Current prevailing data collection paradigms, teleoperation and UMI, demand prohibitive human effort and cost at scale. To scale beyond the limits of manual data collection, we seek to maximize the value of each human demonstration by scalable data generation. To this end, we introduce \method: learning open-\textbf{W}orld mobile m\textbf{AN}ipulation from one demonstration via a synthetic \textbf{DA}ta engine. \method first reconstructs background Gaussian splats and robot-object interaction trajectories from source RGBD observations, as a world substrate for later planning and rendering. It then rearranges contact-rich robot-object interaction segments into extensive spatial configurations, utilizing whole-body motion planning to chain them into new trajectories. To enhance long-horizon robustness, it applies Corrective State Expansion to increase the robot and object state diversity at different stages of mobile manipulation. To unlock cross-environment generalization, trajectories are synthesized on diverse generated 3D worlds from everyday photos. Furthermore, we synthesize photo-realistic observations by compositing rendered robot and object meshes with Gaussian splatting backgrounds. We evaluate our approach on extensive simulation and real-world tasks in various scenes. Experiments show that policies trained with \method achieve long-horizon robustness, broad spatial generalization and cross-environment generalization from one real demonstration. Moreover, \method naturally supports cross-embodiment data generation, validated by zero-shot deployment on another mobile manipulator with a distinct morphology.
\end{abstract}

% \keywords{Robotic Data Generation, Open-World Mobile Manipulation}

% Main paper sections (split into subfiles)
\section{Introduction}
Achieving mobile manipulation in unstructured, open-ended environments, like homes, cafes and offices, has been a long-standing goal in the robotics community. Learning foundation models end-to-end from data has shown great promise in robot manipulation~\citep{physicalintelligence2025pi05,physicalintelligence2026pi07,fang2026molmoact2,kim2024openvla,ye2026world,pertsch2025fast}. However, learning open-world mobile manipulation requires substantial data~\citep{act1} due to its challenging nature: a mobile manipulator has more degrees of freedom and a much larger operation space than tabletop settings. It requires the policy to have strong spatial generalization across diverse configurations~\citep{physicalintelligence2025pi05, yang2025mobipi}, be robust to long-horizon compounding errors~\citep{behavior2025challenge}, and generalize across diverse environments. This necessitates thousands of demonstrations to learn a single task.

Current prevailing data collection paradigms, teleoperation~\citep{cheng2024open, he2024learning,fu2024humanplus} and UMI~\citep{chi2024universal,zhaxizhuoma2025fastumi,xu2026hommi,liu2026rdt2}, require great human effort to collect data at scale. Collecting mobile manipulation data is even more costly. It requires humans to bring physical hardware into diverse environments and repeat the same tasks across different spaces. UMI and human data~\citep{punamiya2026egoverse,kareer2025egomimic,zheng2026egoscale} also require precise odometry and localization techniques to capture accurate poses, which is challenging in the noisy open world.

% another line of synthetic work
Another line of work uses synthetic data to train robot policies~\citep{mandlekar2023mimicgen,jang2025dreamgen,xue2025demogen}. MimicGen~\citep{mandlekar2023mimicgen} and its subsequent extensions~\citep{xue2025demogen,jiang2025dexmimicgen,lin2026humanoidmimicgen,garrett2024skillmimicgen} have proposed to replace the tedious relocate-and-recollect procedure with automatic demonstration generation. MoMaGen~\citep{li2025momagengen} extends it to the mobile manipulation setting via whole-body motion planning, but is still limited to simulation. Due to the sim-to-real gap, they still need real data for sim-to-real co-training~\citep{mandlekar2023mimicgen,lin2026humanoidmimicgen,li2025momagengen} to improve real-world performance. Some work extends the MimicGen-style approach directly to the real-world setting~\citep{xue2025demogen,xu2025r2rgen}. But most of them operate in a sparse point-cloud observation space, which is unsuitable for training large robot foundation models~\citep{physicalintelligence2025pi05,zhen20243d}. Moreover, none of these methods can synthesize diverse demonstrations for open-world mobile manipulation.

In this work, we argue that the potential of a single demonstration is not fully leveraged for policy learning. For mobile manipulation, a single demonstration already contains rich information about the world it acts in. The moving robot's changing views enable consistent 3D scene reconstruction. The robot observations can capture 4D object motions by fusing information across cameras and time. A single demonstration is sufficient to specify the task and to capture the contact-rich interactions essential for success. Learning pixel-to-action policies alone loses this useful information. To maximize the data efficiency of one demonstration, we push the goal to the limit: \emph{\textbf{to learn an end-to-end open-world mobile manipulation policy from a single human demonstration}}.

To this end, we propose \method, learning open-\textbf{W}orld mobile m\textbf{AN}ipulation from one demonstration via a synthetic \textbf{DA}ta engine. \method starts from world reconstruction and generation. For backgrounds, it leverages a sparse-view 2D Gaussian splatting~\citep{huang20242d} method, MAtCha~\citep{guedon2025matcha}, to reconstruct both splats and mesh of the scene. The inputs to MAtCha are the images captured by the head camera during navigation. For foreground objects, it uses BundleSDF~\citep{wen2023bundlesdf} to track the object's 6D pose and reconstruct the mesh from RGBD observations. Inspired by Real2Render2Real~\citep{yu2025real2render2real}, we reconstruct a render-only world without dynamics, mainly for planning and rendering. The reconstructed world serves as a workspace for subsequent trajectory planning and rendering.

To handle the challenges of open-world mobile manipulation, \method adopts several strategies to synthesize novel trajectories. For \textbf{spatial generalization}, the contact-rich robot-object interaction segments are automatically relocated into diverse spatial configurations. Whole-body motion planning is used to chain those segments into complete trajectories. For \textbf{long-horizon robustness}, we propose Corrective State Expansion to simulate policy drifts and increase the robot and object state diversity at different mobile manipulation stages. For \textbf{scene generalization}, we adopt Marble~\citep{marble2026} to scale 3D world generation from everyday photos for novel trajectory synthesis in diverse environments. This makes new scene data generation as simple as taking one picture. We utilize factorized rendering to produce the visual observations. Furthermore, \method naturally supports cross-embodiment data generation, and to our knowledge, demonstrates the first zero-shot \textbf{cross-embodiment generalization} capability for mobile manipulation.

Empirically, we validate \method across simulation and real-world tasks on both VLAs and visuomotor policies. In simulation, \method achieves $\sim$50$\times$ sample efficiency over teleoperated baselines in the single scene setting. In the real world, \method uses one demonstration to achieve 54.8\% average progress score on five long-horizon mobile-manipulation tasks shown in Figure \ref{fig:real_experiments} across 16 different environments, unlocking broad spatial and cross-environment generalization. Moreover, we achieve zero-shot deployment on Linearbot using policies trained solely from synthetic data generated from a single Agibot G1 demonstration, demonstrating cross-embodiment transfer.

\section{Related Work}

\textbf{Data generation for robot learning.}
Collecting teleoperation data is costly, requiring many operators and robot setups~\citep{fu2024mobile, jiang2025behavior}. Simulation-based generation methods~\citep{mandlekar2023mimicgen, garrett2024skillmimicgen, jiang2025dexmimicgen, li2025momagengen} synthesize diverse demonstrations from few demos, but real deployment often requires sim-to-real effort or additional real fine-tuning. DemoGen~\citep{xue2025demogen} avoids dynamics simulation through point-cloud editing and action replanning, but focuses on tabletop spatial generalization rather than long-horizon mobile manipulation. Real2Render2Real~\citep{yu2025real2render2real} relies on manually scanned objects, and 1001Demos~\citep{pan20251001demos} requires pre-scanned scenes for action-view augmentation; both depend on dedicated capture passes that limit scalability, and target tabletop tasks. \method reconstructs scenes and objects directly from the robot demonstration, scaling to long-horizon mobile manipulation in the wild.

\textbf{Imitation learning for mobile manipulation.}
Mobile manipulation is harder than fixed-base manipulation, requiring coordinated base-arm control; teleoperation is correspondingly more expensive~\citep{fu2024mobile, jiang2025behavior}. Mobi-$\pi$~\citep{yang2025mobipi} addresses base-pose selection for pretrained fixed-base policies rather than generating training trajectories. DynaMem~\citep{liu2025dynamem} focuses on modular memory/planning with off-the-shelf grasping rather than end-to-end policy learning. Large vision-language-action policies~\citep{physicalintelligence2025pi05, physicalintelligence2026pi07} provide policy architectures for open-world robotics, but their scaling remains tied to large heterogeneous datasets. \method targets this data bottleneck by turning one mobile-manipulation demonstration into synthetic trajectories across reconstructed and generated scenes.

\section{Method}

\begin{figure}[t]
\centering
\includegraphics[width=\linewidth]{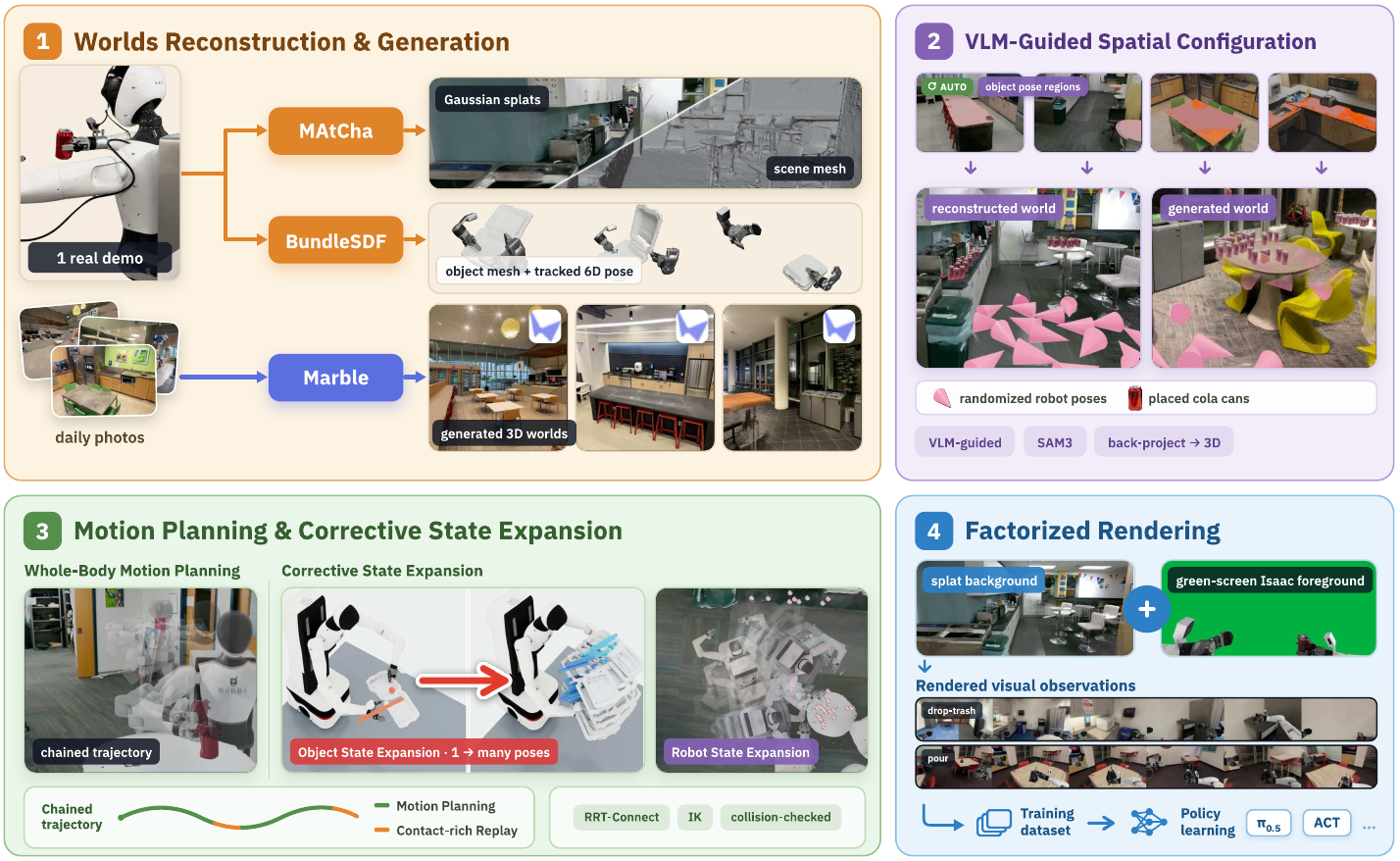}
\caption{\textbf{Pipeline overview of \method.} From a single real demonstration, \method reconstructs and generates 3D worlds, rearranges contact-rich robot-object interaction segments into diverse spatial configurations, chains them into new trajectories with whole-body motion planning, applies Corrective State Expansion to increase robot and object state coverage, and produces photo-realistic observations by factorized rendering.}
\label{fig:pipeline}
\end{figure}

In this section, we propose \method, a data engine that turns one demonstration into diverse mobile manipulation data. As shown in Figure~\ref{fig:pipeline}, we first reconstruct the 4D world from the source RGBD observations and diversify the backgrounds with 3D scenes generated from photos. Then we rearrange contact-rich robot-object interaction segments into extensive spatial configurations in both reconstructed and generated worlds, using whole-body motion planning to chain them into new trajectories. Corrective State Expansion is adopted to increase the robot and object state coverage at different mobile manipulation stages. Finally, we use factorized rendering to achieve fast, photo-realistic rendering of the visual observations for the generated trajectories.

\subsection{Worlds Reconstruction and Generation}

\textbf{3D background scene reconstruction.} To fully leverage the information from one demonstration, \method first performs world reconstruction for both backgrounds and foreground objects from the source observation. This acts as a world substrate for later planning and rendering. For background reconstruction, we first subsample a set of diverse mobile camera views from the source demonstrations. Then we inpaint the foreground from the background using Nano Banana Pro automatically. The background images are fed into a pose-free sparse-view 2D Gaussian Splatting method, MAtCha, to reconstruct both splats and mesh of the scene. To align the reconstructed scene with real scale, we register the scene by aligning the predicted camera height with the corresponding camera height calculated from forward kinematics. This only requires easily accessible robot proprioception and avoids the demand for high-precision SLAM to get accurate camera poses.

\textbf{Robot-object interaction reconstruction.}
We utilize BundleSDF~\citep{wen2023bundlesdf} to simultaneously reconstruct the object mesh and track the 6D pose from RGBD observation streams. Compared with image-to-3D generation methods like SAM3D~\citep{chen2026sam}, this achieves high-fidelity object texture and geometry reconstruction, while eliminating the need to scan each object manually. The reconstruction is also compatible with articulated objects, where it can reconstruct and track the pose for different object parts separately. We provide a full reconstruction running example in Appendix~\ref{app:reconstruction}.
% To put in Appendix: However, during mobile manipulation, sometimes the robot can only observe partial views of the object. Therefore we adopt a generative completion method, that use SAM3D to generate object meshes registered with RGBD observation, and render the unseen faces of the object from real demonstrations. The synthetic view and real view are combined to train BundleSDF for the objects. Therefore, the reconstructed objects achieve high-fidelity texture and geometry at seen faces, while keeping reasonable appearance at unseen faces.

% For intro need to emphasize the world labs scene offer the navigation and manipulation generalization

\begin{figure}[t]
\centering
% Make the image span beyond the text block into the margins (caption stays at text width).
\makebox[\textwidth][c]{\includegraphics[width=1.06\textwidth]{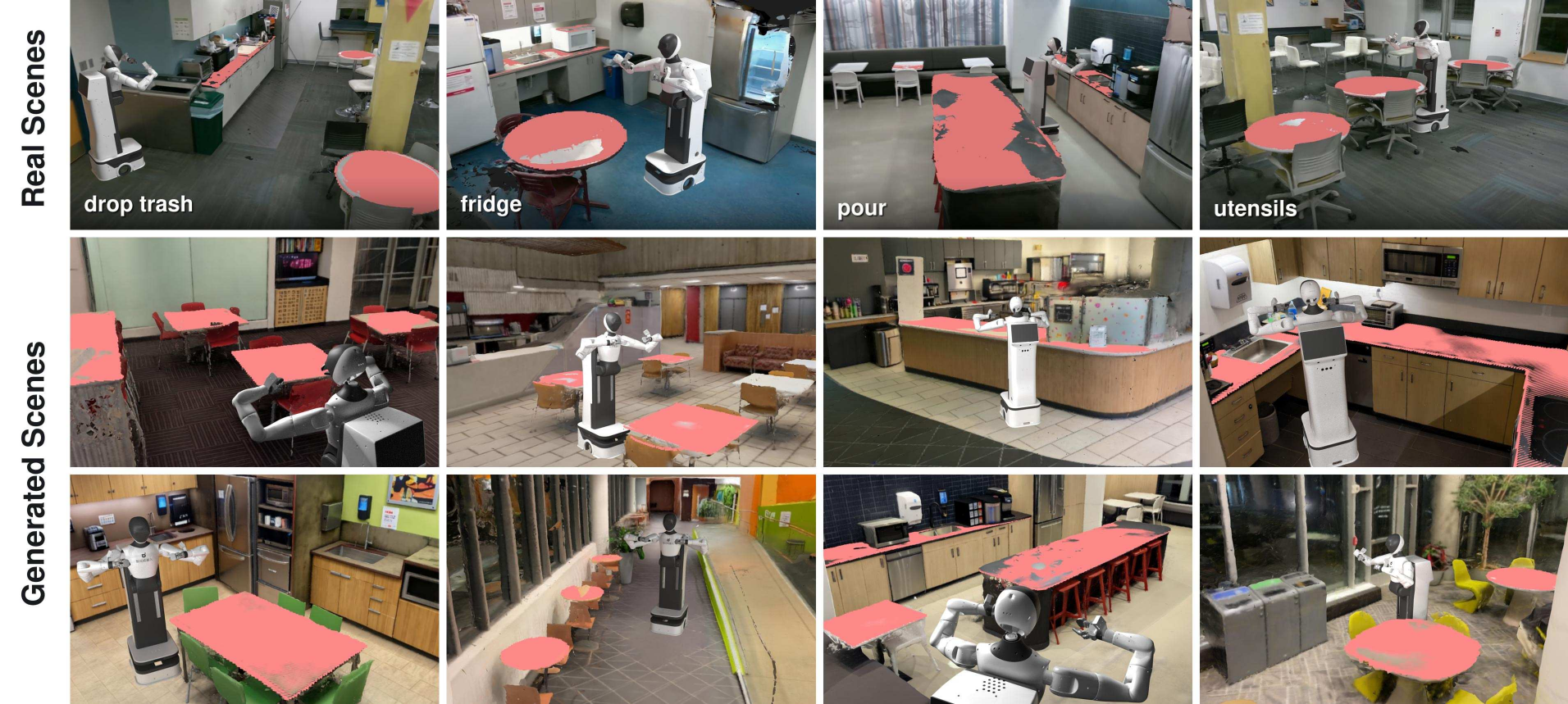}}
\caption{Sampled initial configuration results on reconstructed and generated 3D scenes. Scene and robot meshes are shown in Blender. Red regions indicate the object initial pose sampling regions.}
\label{fig:initial-config}
\vspace{-1em}
\end{figure}

 \textbf{Scalable 3D scene generation.} To enable in-the-wild scene generalization for the policy, we use Marble~\citep{worldlabs2026marble} to generate 3D Gaussian Splats and extract the mesh. We capture only one photo each scene across diverse environments, which are fed into Marble to generate 3D worlds. By augmenting backgrounds with generated 3D worlds, \method scales trajectory generation across diverse scenes to achieve navigation and manipulation generalization in different environments.

\textbf{VLM-guided spatial configuration.} To achieve large spatial coverage, we annotate the contact-rich End Effector-object interaction trajectories and rearrange them into different positions in the scenes. We utilize VLM's commonsense to automatically output the initial randomization regions for replaying the contact-rich manipulation segments. Given one reconstructed or generated scene, we render orbit views of the scene. Then we input the rendered images as well as the manipulation task instruction to prompt Gemini 3.1 Pro to identify entities suitable for initializing the manipulated objects. Then it calls SAM3 to predict the entity masks, which are back-projected to 3D to get the regions. Since Marble-generated scenes often lack metric scale, the VLM leverages commonsense knowledge to infer plausible entity heights, which are then used to recover the scene's metric scale. We showcase some initial configuration results on reconstructed and generated scenes in Figure \ref{fig:initial-config}.
% \vspace{-2em}
\subsection{Novel Trajectory Generation for Mobile Manipulation}
\textbf{Whole-body motion planning.}
After rearranging the contact-rich segments, we use whole-body motion planning to chain them into complete new trajectories. It includes navigation phase and manipulation phase planning. For navigation phases, we first sample a feasible base position and solve whole-body IK that can reach the start EE pose of the contact-rich segments. Then we utilize RRT-Connect~\citep{kuffner2000rrt} to plan a collision-free path for the robot to reach the target base position. For manipulation phases, we utilize RRT-Connect to plan a collision-free arm path to reach the target EE pose from the navigation pose. Then during contact-rich segments, we jointly replay the object tracked pose and the EE motion, with IK applied to solve the arm joint angles. Inspired by MoMaGen, we also apply a visibility cost to keep objects in view during navigation. We detail the full planning procedure in Appendix~\ref{app:trajgen}.
% The planner runs three stages with feasibility evidence passed forward: base navigation by RRT-Connect over SE(2), collision-checked against the scene mesh of the world substrate; whole-body IK to the segment's start state seeded from the nav-selected posture and refined by joint-space RRT-Connect; and replay of the source segment's object-to-end-effector deltas. Nav candidates are scored by a weighted visibility cost jointly with the object's symmetry-free orientation, and the lowest-cost candidate is committed; when line of sight is lost mid-route, the planner inserts a yaw-only spin and replans the remainder from the new yaw. Object placements are pre-sampled offline from VLM-identified support regions in the world substrate, so the planner runs as a pure feasibility search over already-grounded scenes.

\textbf{Corrective State Expansion.} Long-horizon mobile manipulation can easily drift into out-of-distribution situations due to compounding error during navigation and manipulation. Dataset state coverage is crucial for the policy to handle long-horizon drifts. However, plain planning can cause narrow state coverage at different mobile manipulation stages. For example, the navigation planning may always go to the same position relative to the object. But during policy inference, the robot may navigate to a different position relative to the object. To simulate those drifts and increase policy robustness, we propose Corrective State Expansion, including object and robot state expansion.

First we apply \emph{Object State Expansion} to diversify the object state during manipulation. During planning, after the robot reaches the target base position, we perturb the object poses, so that the object position distribution in robot base frame during manipulation has a large spatial coverage. Then we perform IK and arm motion planning again to adapt to the new object poses. This simulates the navigation drift relative to objects and enhances the spatial generalization of manipulation.

% refine to two augmentation? three augmentation tricks to me seems too redundant and complex consider we already has two CSE methods. Also this paragraph is a bit long. I think arm augmentation can be included in Navigation augmentation, they're both randomized in nav phases. we even can not mention the arm augmentation, because I don't think it has much impact.
Second, we perform \emph{Robot State Expansion} to enlarge the robot state coverage during mobile manipulation. Robot proprioceptive state can drift in the long run during manipulation and navigation. Therefore, in the navigation phase, we sample new navigation starts inside a stratified cone behind the original target position, and let the robot turn to the target object and go straight towards it. We also randomly perturb the arm joints during navigation and ease them back to the default carried pose to counter arm drift.
%\emph{Arm Augmentation:} at a random navigation-phase frame, both arms receive bounded Gaussian joint noise with magnitudes graduated proximal-to-distal (small at the shoulder, larger at the wrist), followed by an ease-out recovery to the original joint state; attached objects remain rigidly held by applying the corresponding forward-kinematics delta to their stored pose. This trains the policy to correct arm drift during navigation before it compounds into states outside the training distribution.
% needs to change a term. Augmentation is too ambiguous and not correspond to the corrective state expansion.
In the manipulation phase, we replace the approaching-object segments with a cone-cycloid EE trajectory~\citep{wang2025fieldgen} whose start is sampled in a cone behind the target gripper pose; per-frame IK solves the active arm joint poses. This broadens the state configurations during approaching objects. Together these expansions teach the policy not just on-nominal execution but how to recover from realistic drift accumulated across long horizons.

\textbf{Factorized rendering.}
After novel trajectory generation, we use factorized rendering to get the visual observation. Specifically, we render the background with Gaussian splatting and render the foreground robot and objects with Isaac Sim as a render engine, applying domain randomization on lighting. We then merge the background and foreground to get final frames. This avoids time-consuming large-scene rendering in Isaac Sim and achieves photo-realistic rendering fast.

\section{Experiments}

\begin{figure}[t]
\centering
\includegraphics[width=\linewidth]{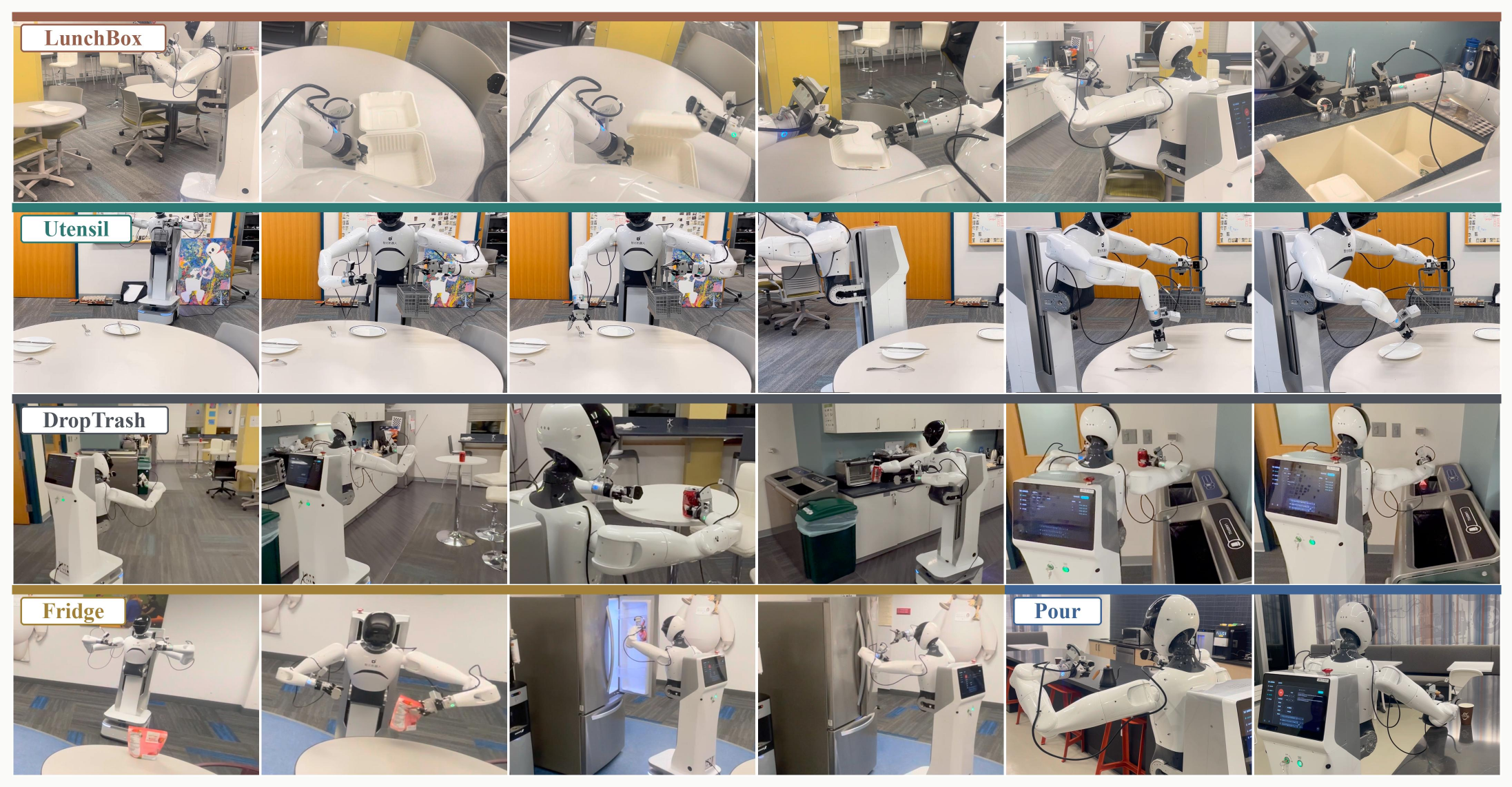}
\caption{\textbf{Real-world tasks.} Five long-horizon mobile-manipulation tasks, each learned from a single real demonstration: \textsc{LunchBox}, \textsc{Utensil}, \textsc{DropTrash}, \textsc{Pour}, and \textsc{Fridge}.}
\label{fig:real_experiments}
\end{figure}

% --- Figure: sim.png (left) + Bigym/BEHAVIOR tables (right) ---
\begin{figure*}[t]
\centering
\small
\renewcommand{\arraystretch}{1.0}

\begin{minipage}[t]{0.48\textwidth}
  \vspace{0pt}
  \centering
  \IfFileExists{figures/sim.png}{%
    \includegraphics[width=\linewidth]{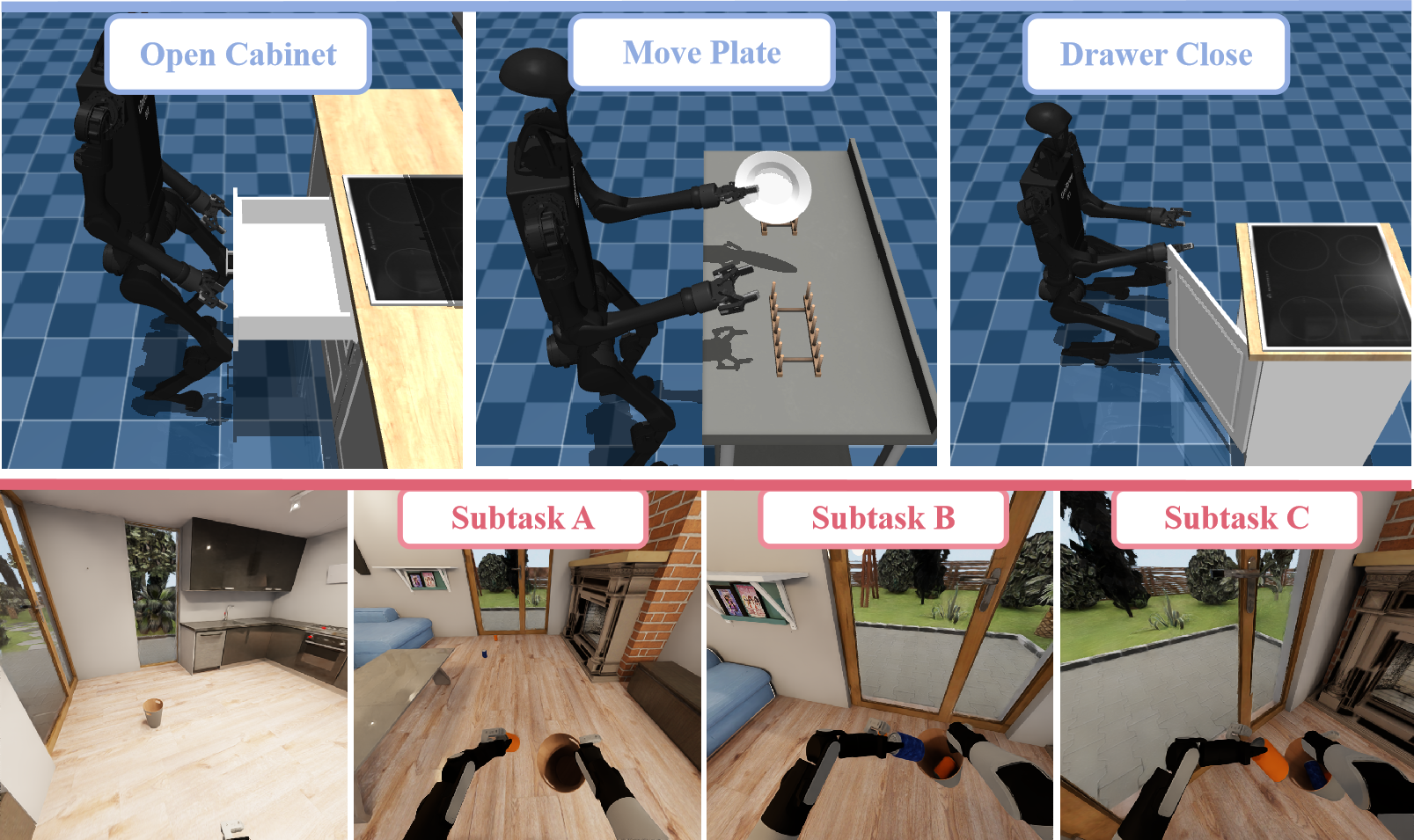}%
  }{%
    \fbox{\parbox[c][0.22\textheight][c]{\linewidth}{\centering Missing file\\\texttt{figures/sim.png}}}%
  }
  \captionof{figure}{\textbf{Demonstration of simulation tasks and subtasks from Bigym~\citep{chernyadev2024bigym} and B1K~\citep{behavior2025challenge}.}}
  \label{fig:sim_tasks}
\end{minipage}\hfill
\begin{minipage}[t]{0.50\textwidth}
  \vspace{0pt}
  \centering
  \captionof{table}{\textbf{Simulation results.}}
  \label{tab:sim_results}
  % Bigym results table snippet (no floating environment here; meant to be \input{} inside a table)
\setlength{\tabcolsep}{3pt}
\begin{tabular}{lcccc}
\toprule
Bigym & \shortstack{\taskname{Saucepan}\\\taskname{to Hob}} & \shortstack{\taskname{Move}\\\taskname{Plate}} & \shortstack{\taskname{Drawer}\\\taskname{Close}} & \textbf{Avg} \\
\midrule
\method (Ours)   & \hl{86.7} & \hl{40.0} & \hl{100.0} & \hl{75.6} \\
full source demo & 86.0 & 48.0 & 100.0 & 78.0 \\
\bottomrule
\end{tabular}

  \vspace{4pt}

  % BEHAVIOR results table snippet (no floating environment here; meant to be \input{} inside a table)
\setlength{\tabcolsep}{4pt}
\begin{tabular}{lcccc}
\toprule
BEHAVIOR & \taskname{A} & \taskname{B} & \taskname{C} & \textbf{Q-score} \\
\midrule
\method   & \hl{20.0} & \hl{16.7} & \hl{13.3} & \hl{16.67} \\
20 demos  & 6.7  & 3.3  & 0.0  & 3.33 \\
50 demos  & 16.7 & 13.3 & 3.3  & 11.11 \\
\textcolor{gray!50}{200 demos} & \textcolor{gray!50}{80.0} & \textcolor{gray!50}{76.7} & \textcolor{gray!50}{30.0} & \textcolor{gray!50}{62.22} \\
\bottomrule
\end{tabular}

\end{minipage}
\end{figure*}

% policy learning?
% A sim experiments: two benchmark, two policy learning
% B real experiments setup: 1. main experiment: progress score in appendix;  baseline: CSE; 
% 2. Scaling curve: generalization across scene; cola scaling. render cost
% C Case study: visibility. 
% D Visualization of CSE; Visualization of generated demos in domain and world labs

\subsection{Simulation Experiments}

We evaluate \method across different policy architectures and simulation benchmarks to validate the general effectiveness. For all the simulation experiments, we only utilize the RGBD observation and joint state information from the demonstrations. No privileged information is leveraged to align with the real-world setting. 

\textbf{Bigym.} First, we choose three mobile manipulation tasks from Bigym~\citep{chernyadev2024bigym} shown in Figure~\ref{fig:sim_tasks} and generate 100 demonstrations from one source demonstration. Then we choose ACT for policy learning, to validate the effectiveness of \method on small policies. Baselines are ACTs trained on the full source dataset, around 40--60 demos per task. We report success rate over 50 rollouts.

\textbf{BEHAVIOR Challenge.} We choose a three-minute-long task from BEHAVIOR Challenge~\citep{behavior2025challenge}, which contains four navigation stages and four manipulation subtasks shown in Figure~\ref{fig:sim_tasks} to stress-test our method on long-horizon mobile manipulation. We perform full-parameter fine-tuning of $\pi_{0.5}$~\citep{physicalintelligence2025pi05} on the datasets. To investigate the data efficiency, we generate 1360 demonstrations from one official demonstration and compare with $\pi_{0.5}$ trained on 20, 50, and 200 official demonstrations separately. We report the Q-score (progress score defined by official implementation) over 30 evaluation episodes.

\textbf{Results.}
The simulation evaluation results are presented in Table~\ref{tab:sim_results}. For Bigym, using data generated from one source demonstration, \method achieves comparable results to baselines trained on $\sim$50 demonstrations. This indicates the effectiveness of our method for localized mobile manipulation. Even with limited viewpoint changes during movement, our pipeline still reliably reconstructs the world and generates high-fidelity demonstrations. For BEHAVIOR Challenge, our approach even outperforms those trained on 20 and 50 source demonstrations. This validates our pipeline's effectiveness for generating high-quality trajectories for extremely long-horizon mobile manipulation. Together, the simulation results show that \method achieves $\sim$50$\times$ data efficiency, reducing human effort for mobile manipulation data collection.
\subsection{Real-World Experiments} % including ablation?

\begin{table}[t]
\centering
\caption{\textbf{Real-world mobile manipulation results on Agibot G1.} Progress scores are mean completed-subtask percentages over 10 trials; Average is the unweighted task mean. The Utensil ablation retains its legacy two-subtask score.}
\label{tab:realworld_results}
\small
\renewcommand{\arraystretch}{1.12}
% Real-world results table snippet (no floating environment here; meant to be \input{} inside a table)
% Original wide layout: methods are rows and all tasks appear on one line.
\begin{tabular*}{\linewidth}{@{\extracolsep{\fill}}lcccccc@{}}
\toprule
Method & \taskname{Lunch Box} & \taskname{Utensil} & \taskname{Drop Trash} & \taskname{Fridge} & \taskname{Pour} & \textbf{Average} \\
\midrule
1 real demo & 0.0\% & 0.0\% & 0.0\% & 0.0\% & 0.0\% & 0.0\% \\
WANDA (Ours) & \hl{55.0\%} & \hl{52.5\%} & \hl{75.0\%} & \hl{46.7\%} & \hl{45.0\%} & \hl{54.8\%} \\
WANDA w/o CSE & 10.0\% & 15.0\% & 35.0\% & 13.3\% & 5.0\% & 15.7\% \\
\bottomrule
\end{tabular*}

\end{table}

\begin{figure}[t]
  \centering
  \includegraphics[width=0.95\linewidth]{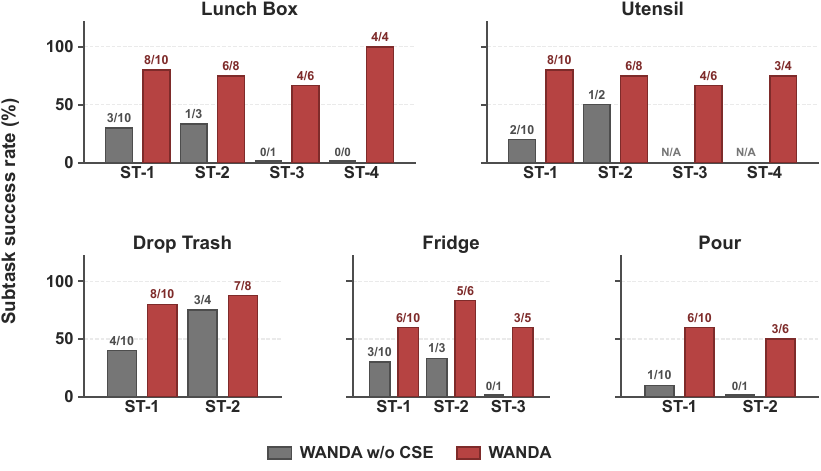}
  \caption{\textbf{Per-subtask success rate on the real-world tasks.} Conditional success with raw counts; $0/0$ means never reached and N/A denotes stages unrecorded in the legacy Utensil ablation.}
  \label{fig:subtask_sr}
\end{figure}

We design five mobile manipulation tasks on Agibot G1, as illustrated in Figure~\ref{fig:real_experiments}. These tasks emphasize long horizon robustness and spatial generalization over different regions in the scenes:
\begin{list}{\textbullet}{%
  \setlength{\leftmargin}{1.4em}% move list left
  \setlength{\labelsep}{0.5em}% space between bullet and text
}
  \item \textit{Lunch Box:} The robot needs to first navigate to the lunch box on different tables, fold it, then take it to the kitchen sink and place it in the sink. We generate 396 trajectories for this task.
  \item \textit{Utensil:} The robot is required to first search the knife on the plate, navigate to it, pick up the knife precisely and place it in the storage basket. 510 generated trajectories are used for training.
  \item \textit{Drop Trash:} The robot needs to first find the cola trash over different tables in the rooms, grasp it and navigate to the trash can and throw the trash. We generate 407 trajectories for this task.
  \item \textit{Pour:} The robot needs to navigate to the teapot randomized in the large kitchen table, take the teapot to the cup and pour into it. 740 trajectories are generated for the task.
  \item \textit{Fridge:} The robot is required to take the grocery bag randomized on the round table, store it in the fridge and close the fridge door. We utilize 263 generated demonstrations to train the policy.

\end{list}
We use PICO VR to teleoperate Agibot G1 to collect one source demonstration for each task. We perform full-parameter fine-tuning on $\pi_{0.5}$ on the generated data. We compare the full \method with a variant that removes Corrective State Expansion (CSE) to assess the necessity of this component.

\textbf{Metric.} We conduct 10 trials for each task and report the progress score (the percentage of completed subtasks). The progress score for each task is defined in Appendix~\ref{app:progress}.

\textbf{Results.} The results are presented in Table~\ref{tab:realworld_results}, with per-subtask success rates in Figure~\ref{fig:subtask_sr} and subtask definitions in Appendix~\ref{app:progress}. Fine-tuning $\pi_{0.5}$ on only the single real demonstration (\emph{without} our data generation) yields a 0 progress score on every task, i.e., it fails to reliably complete even the first subtask; we include this ``1 real demo'' baseline in Table~\ref{tab:realworld_results}. In contrast, \method produces meaningful progress on all five long-horizon real-world mobile manipulation tasks from a single human demonstration. For \textit{Lunch Box}, the robot must complete a multi-stage task lasting around three minutes; \method remains robust to long-horizon drift and completes the full sequence. For \textit{Drop Trash}, the robot steadily navigates to the trash placed on different tables, grasps it, and carries it to the trash can. This indicates the effectiveness of our method in achieving spatial generalization. For \textit{Utensil}, the robot navigates a long distance to reach the target base position and grasps the knife precisely. Together, these tasks show that our method is compatible with both rigid and articulated objects.

\textbf{CSE Ablation.} The performance of the non-CSE variant degrades significantly compared with full \method. \method without CSE is vulnerable to compounding state deviations. Before manipulation, the robot may reach a pose with noticeable error relative to the source demonstration, leading to unseen robot-object configurations that are not covered by non-CSE data. During manipulation, small drifts in the arm joint state can further push the policy out of distribution when approaching the object. In contrast, \method remains robust under such deviations, demonstrating the importance of Corrective State Expansion. Figure~\ref{fig:cse_vis} visualizes how CSE mitigates these deviations.

\begin{figure}
  \centering
  \captionsetup{skip=2pt}
  \begin{minipage}[t]{0.34\linewidth}
    \centering
    \IfFileExists{figures/scaling_curve.pdf}{%
      \includegraphics[width=\linewidth]{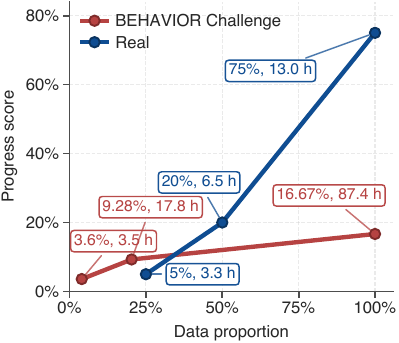}%
    }{%
      \fbox{\parbox[c][0.16\textheight][c]{\linewidth}{\centering Missing file\\\texttt{figures/scaling\_curve.pdf}}}%
    }\\[-2pt]
    {\small\textbf{(a)} Data proportion scaling}
  \end{minipage}\hfill
  \begin{minipage}[t]{0.64\linewidth}
    \centering
    \includegraphics[width=\linewidth]{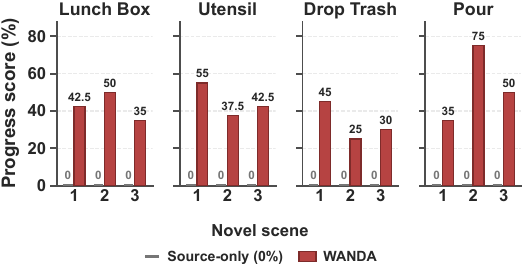}\\[-2pt]
    {\small\textbf{(b)} Progress score on novel scenes}
  \end{minipage}

  \caption{\textbf{Scalability results.} (a) Performance versus generated-data proportion. (b) Per-task progress scores in three novel scenes; each task--scene pair compares \method with the 0\% source-only baseline.}
  \label{fig:scaling}
\end{figure}

\begin{figure}
  \centering
  \includegraphics[width=\linewidth]{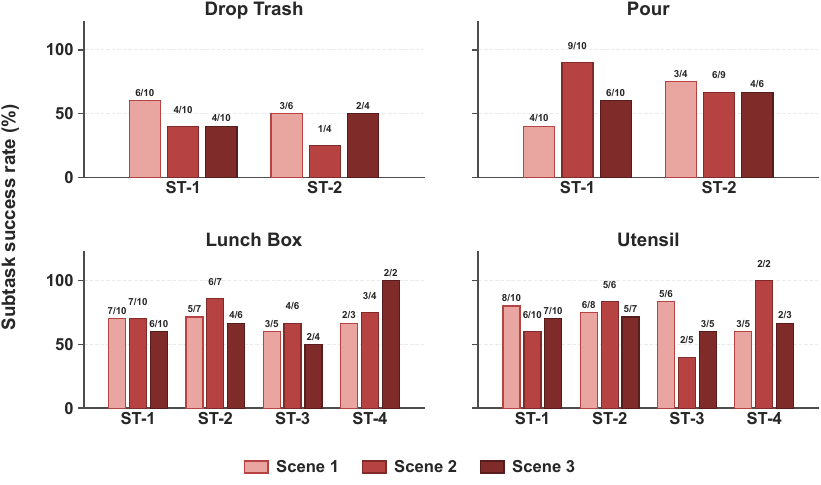}
  \caption{\textbf{Per-subtask success rate across novel scenes.} \method is trained on Marble-generated worlds and evaluated on Drop Trash, Pour, Lunch Box, and Utensil in three real scenes; we use the same conditional metric as Figure~\ref{fig:subtask_sr}. Subtask definitions are given in Appendix~\ref{app:progress}.}
  \label{fig:scenegen_subtask}
\end{figure}

\subsection{Scalability Study} % including scene scaling and in domain demo scaling
To assess the scalability of the proposed data generation pipeline, we evaluate our methods on two scaling dimensions: scaling in-domain demonstrations in the source scene, and scaling generated scene diversity in the generated demonstrations.

\textbf{Scaling generated demonstrations for in-domain improvement.}
We perform this experiment both in BEHAVIOR Challenge and in real-world \textit{Drop Trash} task. We sample subsets of the full generated dataset and perform $\pi_{0.5}$ fine-tuning under the same training configurations. As shown in Figure~\ref{fig:scaling}(a), scaling the generated data achieves steady improvement in both simulation and real world. This implies the scalability of \method for in-domain performance improvement.

\textbf{Scaling generated scenes for open-world generalization.}
To validate the scene generalization ability of our approach, we test \method on different environments. For each scene, we take a single photo and feed it into Marble to generate a corresponding 3D world. We train the policy on data synthesized from both source scenes and these photo-generated worlds, and evaluate it in the same real-world scenes where the photos were taken. Although the generated worlds are derived from these environments, they still contain substantial visual and geometric discrepancies from reality. Consequently, successful task execution requires strong scene generalization to bridge this gap.

During evaluation, we roll out the policy trained on only the source scene and the policy trained on generated scenes in each environment for 10 episodes and report the progress score. As shown in Figure~\ref{fig:scaling}(b), source-only baseline stays near a zero progress score on every novel scene, whereas \method sustains substantial progress across all four tasks and three novel scenes. Figure~\ref{fig:scenegen_subtask} reports the per-subtask breakdown across the three scenes. This shows \method scales scene diversity, making open-world data generation as easy as taking one photo.

\begin{table}[t]
\centering
\caption{\textbf{Data-generation time cost.} Wall-clock hours to generate one hour of 30\,fps demonstration data.}
\label{tab:cost_compare}
\small
% Data-generation time-cost table (meant to be \input{} inside a table)
\setlength{\tabcolsep}{7pt}
\begin{tabular}{lccccc}
\toprule
 & Human teleop & \multicolumn{2}{c}{Ours} & \multicolumn{2}{c}{MoMaGen} \\
\cmidrule(lr){2-2}\cmidrule(lr){3-4}\cmidrule(lr){5-6}
 & 1 teleoperator & 1 GPU & 8 GPUs & 1 GPU & 8 GPUs \\
\midrule
Time to generate 1\,h of data (h) & 1.5 & \hl{3.0} & \hl{0.38} & 19.4 & 2.4 \\
\bottomrule
\end{tabular}

\end{table}

\textbf{Data Generation Cost.} Our method is scalable with the compute to generate data. Table~\ref{tab:cost_compare} compares the wall-clock time to generate one hour of 30\,fps demonstration data. Through factorized rendering, \method generates one data-hour with about 3 GPU-hours, 6.4$\times$ less than MoMaGen~\citep{li2025momagengen} at equal GPU count; on a single 8$\times$L40S node this is only 0.38 wall-clock hours, about 4$\times$ faster than a human teleoperator. Crucially, \method needs only one human demonstration per task upfront, whose labor amortizes toward zero as the generated dataset scales, whereas teleoperation incurs roughly 1.5 human-hours for every data-hour, estimating the average environment reset time to be half the demonstration time. A per-task cost breakdown is provided in Appendix~\ref{app:cost}.

% \vspace{-1em}
\subsection[Zero-shot Cross-Embodiment Generalization]{Zero-shot Cross-Embodiment\\Generalization}
\label{sec:cross_emb}
\method naturally supports cross-embodiment data generation, since the reconstructed world is embodiment-agnostic for planning and rendering. Also, a single demonstration is sufficient to specify the task and capture the contact-rich interactions essential for success, which is also embodiment-agnostic for most non-prehensile manipulation tasks.

\begin{wrapfigure}{r}{0.48\textwidth}
  \vspace{-0.7em}
  \centering
  \captionsetup{skip=3pt}
  \includegraphics[width=\linewidth]{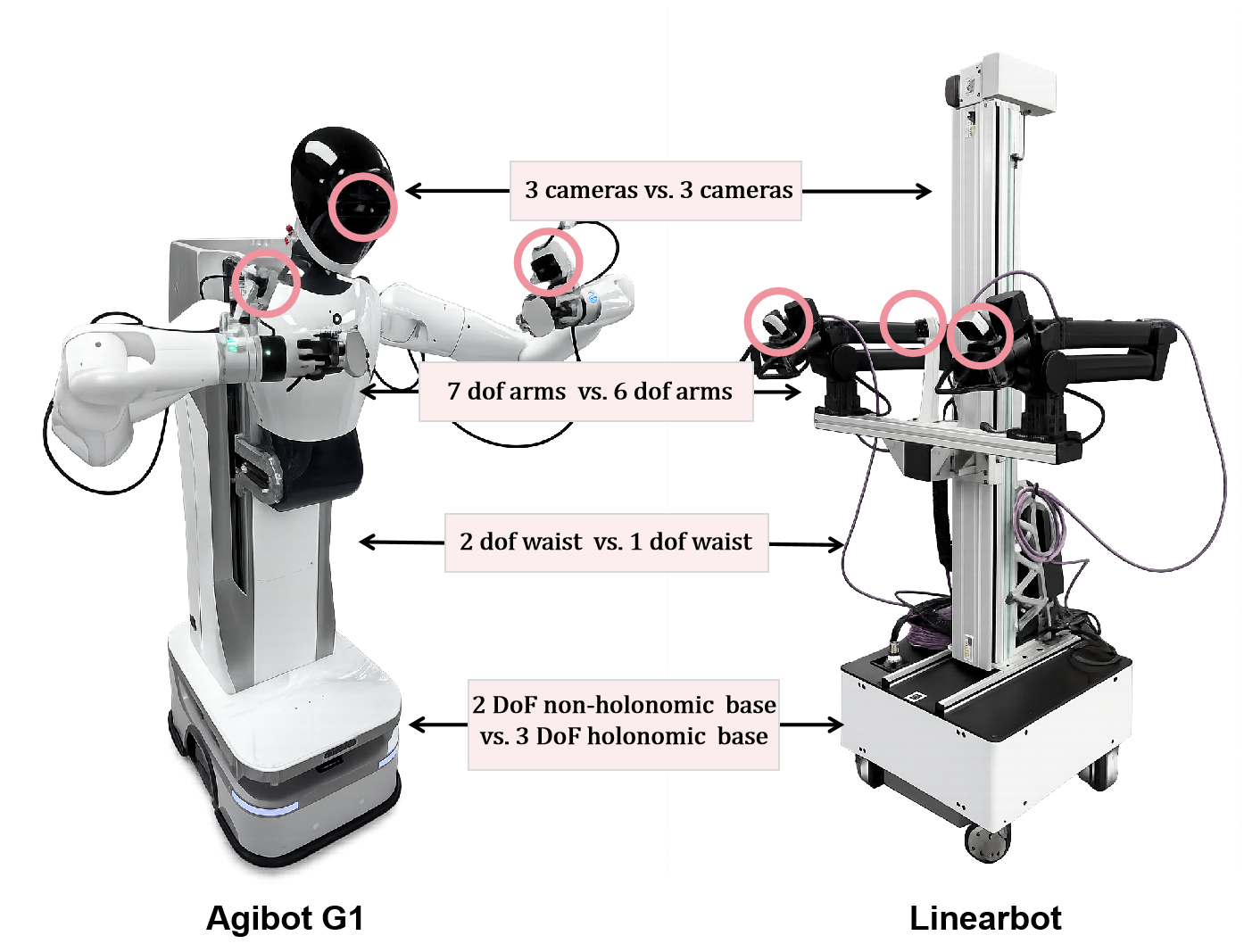}
  \caption{Cross-embodiment data generation from Agibot G1 to Linearbot.}
  \label{fig:cross_emb}

  \vspace{0.4em}
  \captionof{table}{Linearbot subtask success rate.}
  \label{tab:cross_emb_sr}
  \footnotesize
  \setlength{\tabcolsep}{5pt}
  \renewcommand{\arraystretch}{1.15}
  \begin{tabular}{@{}lc@{}}
    \toprule
    Subtask & Success rate \\
    \midrule
    Grasp the cola can & 5/10 \\
    Throw into the trash bin & 3/5 \\
    \bottomrule
  \end{tabular}
  \vspace{-0.6em}
\end{wrapfigure}

Here we perform cross-embodiment data generation from Agibot G1 to Linearbot, shown in Figure~\ref{fig:cross_emb}. These two robots have large morphology gaps and different camera placements. We choose \textit{Drop Trash} for validation. 400 Linearbot demonstrations are generated from one Agibot G1 real demonstration and fed into $\pi_{0.5}$ training. The policy is then zero-shot deployed on a physical Linearbot. Over ten rollouts it reaches a $40\%$ progress score (per-subtask success rates in Table~\ref{tab:cross_emb_sr}); the generated observations are shown in Appendix~\ref{app:cross_emb}. To our knowledge, this demonstrates the first zero-shot cross-embodiment generalization capability in mobile manipulation.

\FloatBarrier
\subsection{Visualization}

\textbf{Generated demonstration visualization.} We show the visualization of head camera views from the synthetic demonstrations in Figure~\ref{fig:teaser}. For all generated camera streams including wrist cameras, please refer to Appendix~\ref{app:videos}. Our approach generates photo-realistic observations for the source scenes and diverse generated scenes, providing high-quality data for downstream policy learning.

\begin{figure}[H]
  \centering
  \vspace{-0.5em}
  \captionsetup{skip=2pt}
  \captionsetup[subfigure]{skip=2pt}
  \begin{subfigure}[t]{0.47\linewidth}
    \centering
    \IfFileExists{figures/lunch_box_base_frame_offset_comparison.pdf}{%
      \includegraphics[width=\linewidth]{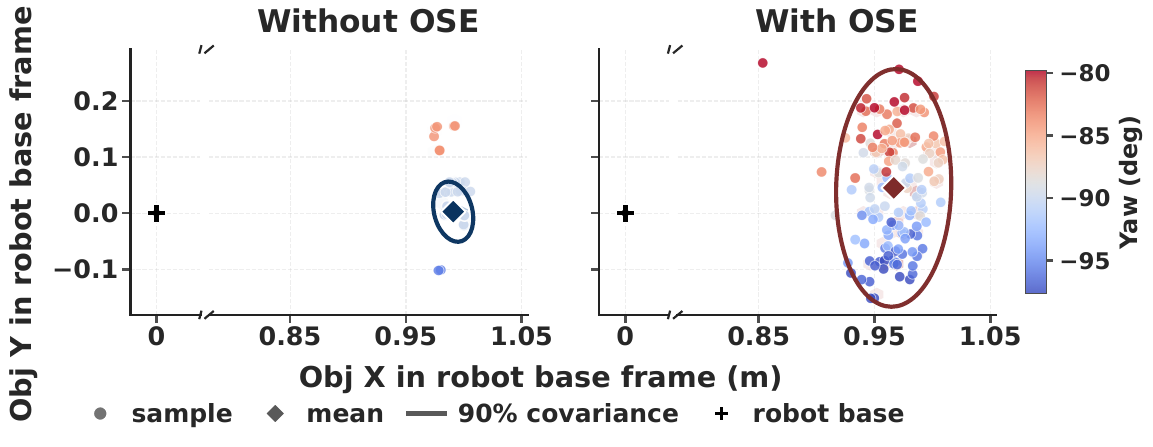}%
    }{%
      \fbox{\parbox[c][0.14\textheight][c]{\linewidth}{\centering Missing file\\\texttt{figures/lunch\_box\_base\_frame\_offset\_comparison.pdf}}}%
    }
    \caption{Object state expansion (OSE)}
  \end{subfigure}\hfill
  \begin{subfigure}[t]{0.47\linewidth}
    \centering
    \IfFileExists{figures/fieldgen_start_pose_cone_distribution.pdf}{%
      \includegraphics[width=\linewidth]{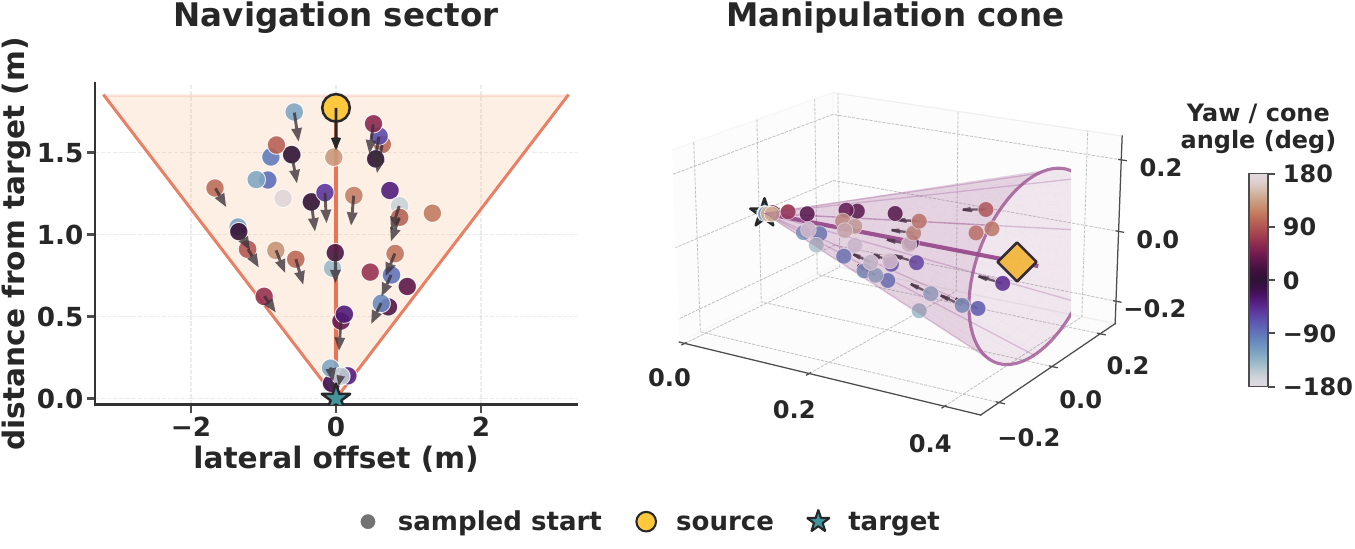}%
    }{%
      \fbox{\parbox[c][0.14\textheight][c]{\linewidth}{\centering Missing file\\\texttt{figures/fieldgen\_start\_pose\_cone\_distribution.pdf}}}%
    }
    \caption{Robot state expansion}
  \end{subfigure}

  \caption{\textbf{Corrective State Expansion.}}
  \label{fig:cse_vis}
  \vspace{-0.5em}
\end{figure}

\textbf{Corrective State Expansion visualization.} To understand the role of Corrective State Expansion, we visualize the object position distribution in robot base frame during manipulation and the robot state distribution during navigation and manipulation from a subset of the dataset. As shown in Figure~\ref{fig:cse_vis}(a), CSE achieves a much larger spatial distribution of object positions during manipulation than the non-CSE version, which enhances the manipulation spatial robustness. Figure~\ref{fig:cse_vis}(b) shows the robot pose distributions over the target. CSE greatly enhances the density and diversity of robot poses for the target, making it able to recover from pose drift.  

% \vspace{-1em}

% (Removed Table 2: ablations table)

\section{Limitations and Future Work}

\method has two main limitations. First, it handles only rigid and articulated objects. Reconstruction tracks the 6D poses of rigid parts and replays them, so soft and deformable objects, such as cloth and fluids, cannot be reconstructed or generated; in such cases \method replays the demonstrated motion rather than modeling the underlying deformation. Second, as a multi-stage data engine, errors can compound across stages: a reconstruction or planning failure in one stage propagates to the subsequent navigation and manipulation stages, so generating long-horizon trajectories occasionally requires human effort to intervene and correct failed stages.

These limitations point to several directions for future work. To relax the rigid-object assumption, the reconstruction stage could be extended with 4D Gaussian Splatting~\citep{wu20244d} to capture complex, deformable object interactions. Because evaluating open-world mobile manipulation is time-consuming and stochastic---it requires physically relocating the robot across many sites---incorporating physics into \method and extending it toward real-to-sim evaluation is a promising route to benchmarking mobile manipulation more efficiently and reproducibly. Finally, co-training the synthesized data with UMI data and human videos is an interesting avenue for further broadening the diversity and coverage of the learned policies.

\section{Acknowledgments}
This work was partially funded by NSF Award \#2512805. Research supported by the NVIDIA Academic Grant Program using NVIDIA RTX PRO 6000 Blackwell Max-Q GPUs and the NVIDIA Jetson AGX Thor platform.
Guanya Shi holds concurrent appointments as an Assistant
Professor at Carnegie Mellon University and as an Amazon
Scholar. This paper describes work performed at Carnegie
Mellon University and is not associated with Amazon.
We greatly thank AppLovin for providing us with the compute resources for this project. We thank Tairan He, Chaoyi Pan and Junfeng Ni for their insightful discussions and feedback. We also thank Zeji Yi, Yi Yang and Angchen Xie for their support with the experiments.

\clearpage
\bibliographystyle{plainnat}
\bibliography{wodem_refs}

\clearpage
\tableofcontents

\clearpage
\appendix

% Figure-heavy appendix: let pages end naturally instead of stretching
% vertical glue into mid-page gaps (\flushbottom underfull pages).
\raggedbottom

% =====================================================================
% Appendix outline. Sections are ordered by importance; the order is
% fixed because the main text cross-references these labels. Fill in
% content under each heading WITHOUT changing the section order.
% =====================================================================

\section{Reconstruction Running Example}
\label{app:reconstruction}
% TODO: a running example from raw observations to the reconstructed
% scene and objects (figure walking through the full pipeline).

\subsection{Demonstration Post-process Interface}

\begin{figure}[H]
  \centering
  \includegraphics[width=\linewidth,height=0.27\textheight,keepaspectratio]{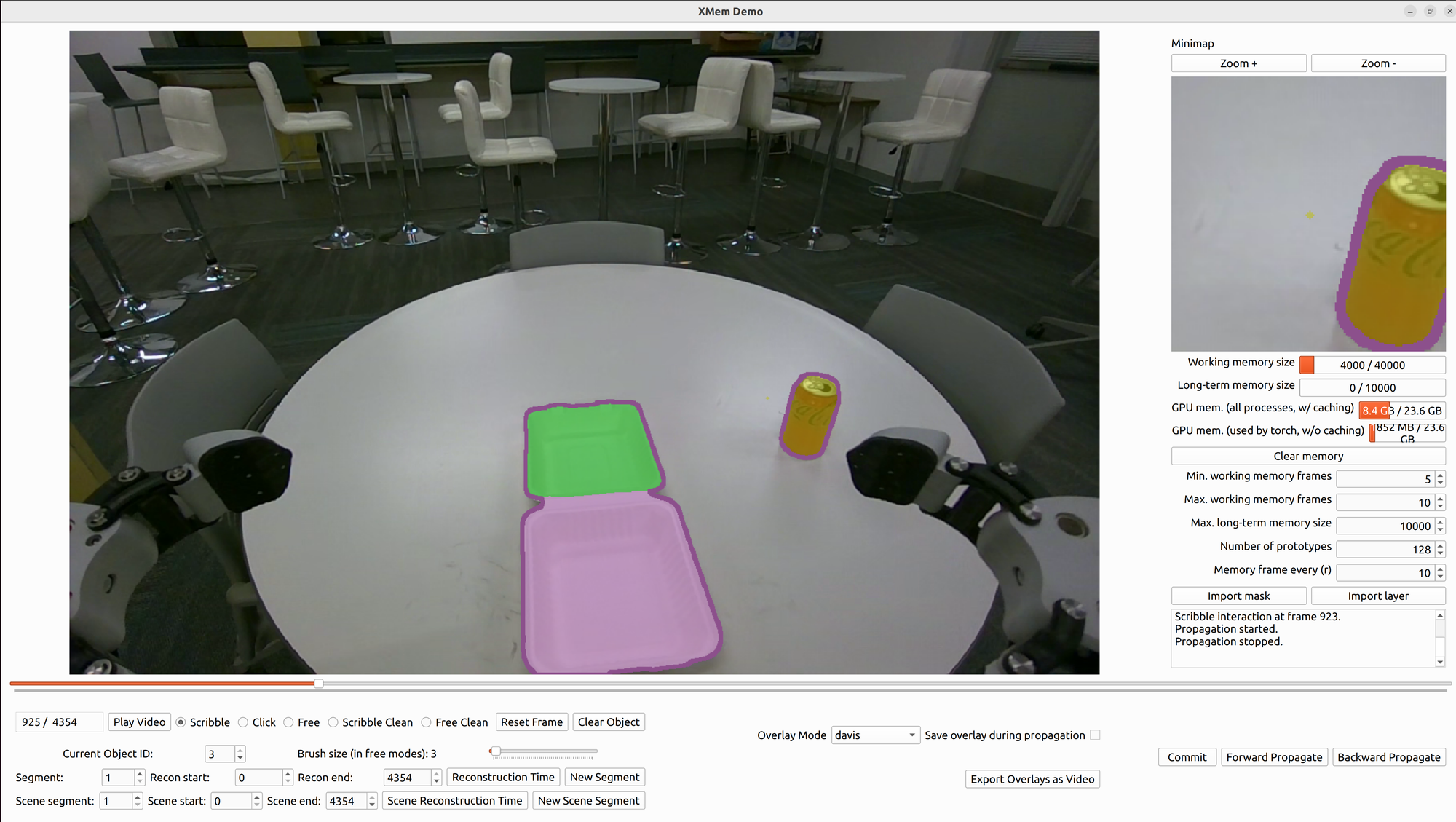}
  \caption{Demonstration post-process interface used to annotate contact-rich and navigation segments.}
  \label{fig:annotation_ui}
\end{figure}

Given one collected real demonstration, we first segment the foreground object masks, annotate the contact-rich segments and navigation segments via a user-friendly interface (Figure~\ref{fig:annotation_ui}). We utilize XMem~\citep{cheng2022xmem}, a long-term video segmentation approach, to get the foreground object masks along the video. For background reconstruction, we annotate the navigation segments, and fill the object mask region with green-screen color. Then we prompt Nano Banana Pro (Gemini 3 Pro Image) to remove the foreground robot arms and object masks and inpaint the background, which are then fed into background reconstruction pipeline. The prompt is shown in Appendix \ref{inpaint_prompt}. For foreground reconstruction, we annotate the contact-rich manipulation segments to get corresponding object masks. Then the masks with corresponding RGB and Depth images are then fed into BundleSDF~\citep{wen2023bundlesdf} for foreground reconstruction. The annotation process only requires the user to click the segmented object a few times. Together with the segmentation model's propagation time, annotation costs $5$ to $30$ minutes, varying among tasks. This process could even be replaced by an automatic agent workflow in the future to remove the human annotation effort.

% Reconstruction results figure: placed before A.2 to keep it on the first appendix page.
\begin{figure}[H]
  \centering
  \includegraphics[width=\linewidth,height=0.29\textheight,keepaspectratio]{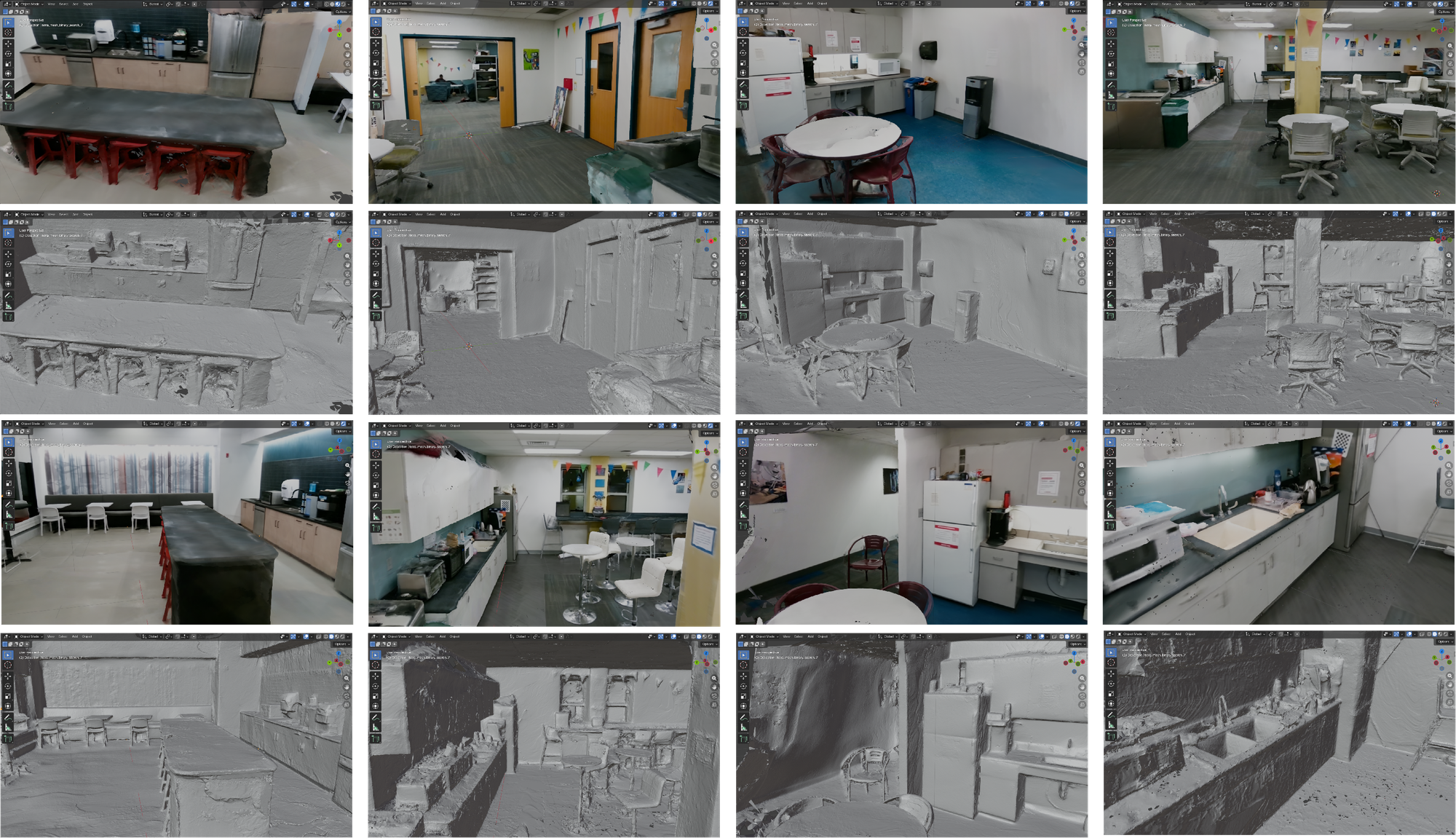}
  \caption{Background reconstruction results (textured mesh and geometry mesh) visualized in Blender. From left to right: \textit{Pour}, \textit{Drop Trash}, \textit{Fridge}, \textit{Utensil}.}
  \label{fig:reconstruction_results}
\end{figure}

\subsection{Background Reconstruction}
\label{app:recon_bg}

Given inpainted background images, we utilize a pose-free sparse-view 2D Gaussian splatting method, MAtCha~\citep{guedon2025matcha} to reconstruct both splats and geometry meshes, and predict camera poses in the world frame. To align the scale and orientation of the reconstructed splats and meshes, we perform registration to minimize the predicted camera pose $z$ and the actual camera height calculated via forward kinematics from recorded joint states. This avoids the demand on high-precision odometry or SLAM, making the pipeline robust to noisy real-world observation. More background reconstruction results are shown in Figure~\ref{fig:reconstruction_results}, where we show the textured mesh and geometry mesh in Blender.

% TODO: background (scene) reconstruction details.

\subsection{Foreground Reconstruction}

\begin{figure}[H]
  \centering
  \includegraphics[width=\linewidth]{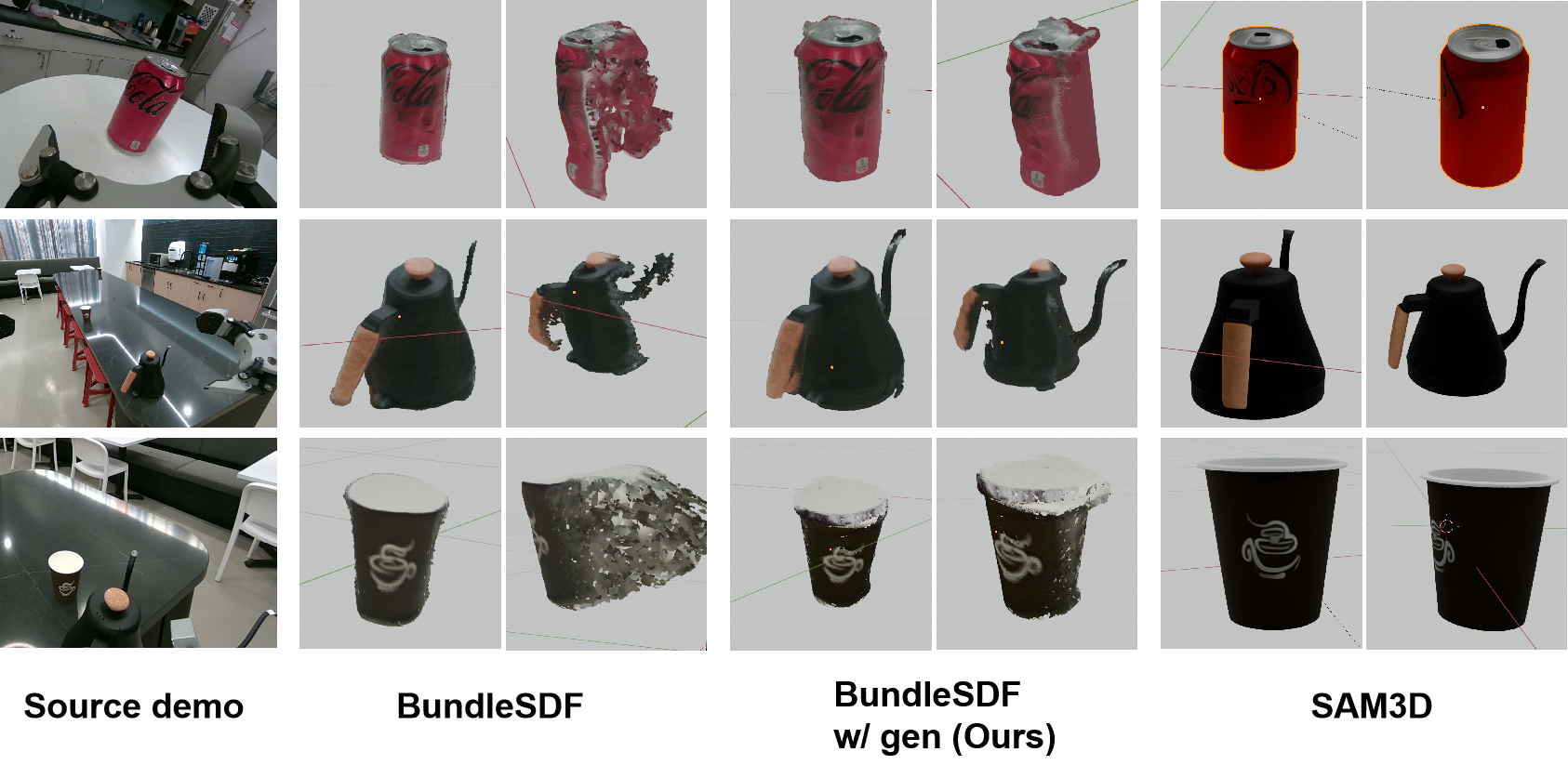}
  \caption{Foreground reconstruction: BundleSDF vs. BundleSDF with generative completion.}
  \label{fig:bundlesdf_gen_exp}
\end{figure}

We adopt BundleSDF~\citep{wen2023bundlesdf}, a SDF~\citep{sitzmann2020implicit} and NeRF~\citep{mildenhall2021nerf} based approach to jointly reconstruct the object geometry and texture and track the object 6D poses. It accepts RGB, object segmentation mask and depth image streams from the observation streams of the contact-rich segments. We adjust the parameters of BundleSDF to make it suitable for fine-grained texture reconstruction. Besides, sometimes the source demonstration cannot have a complete viewpoint coverage over the objects, therefore the reconstruction can be incomplete. For such cases, we adopt a generative completion method inspired by UA-Pose~\citep{li2025ua}. Specifically, we utilize SAM3D~\citep{chen2026sam} to generate 3D mesh from single image of the object from the demonstration. Although SAM3D-generated mesh fails to reconstruct the photo-realistic texture, it provides a good geometry shape prior for the non-seen part of objects. Therefore, we first reconstruct the raw object mesh from incomplete real RGBD observation and compute the visibility of each reconstructed face. Then we perform registration to align the SAM3D generated mesh scale and 6D pose with the reconstructed mesh, then render the images from SAM3D generated mesh on those viewpoints unseen in real observation. Then we fuse the real RGBD images and synthetic images rendered by SAM3D mesh, and train the object SDF and NeRF to get final textured meshes.

\begin{figure}[H]
  \centering
  \includegraphics[width=\linewidth,trim=3mm 0mm 3mm 0mm,clip]{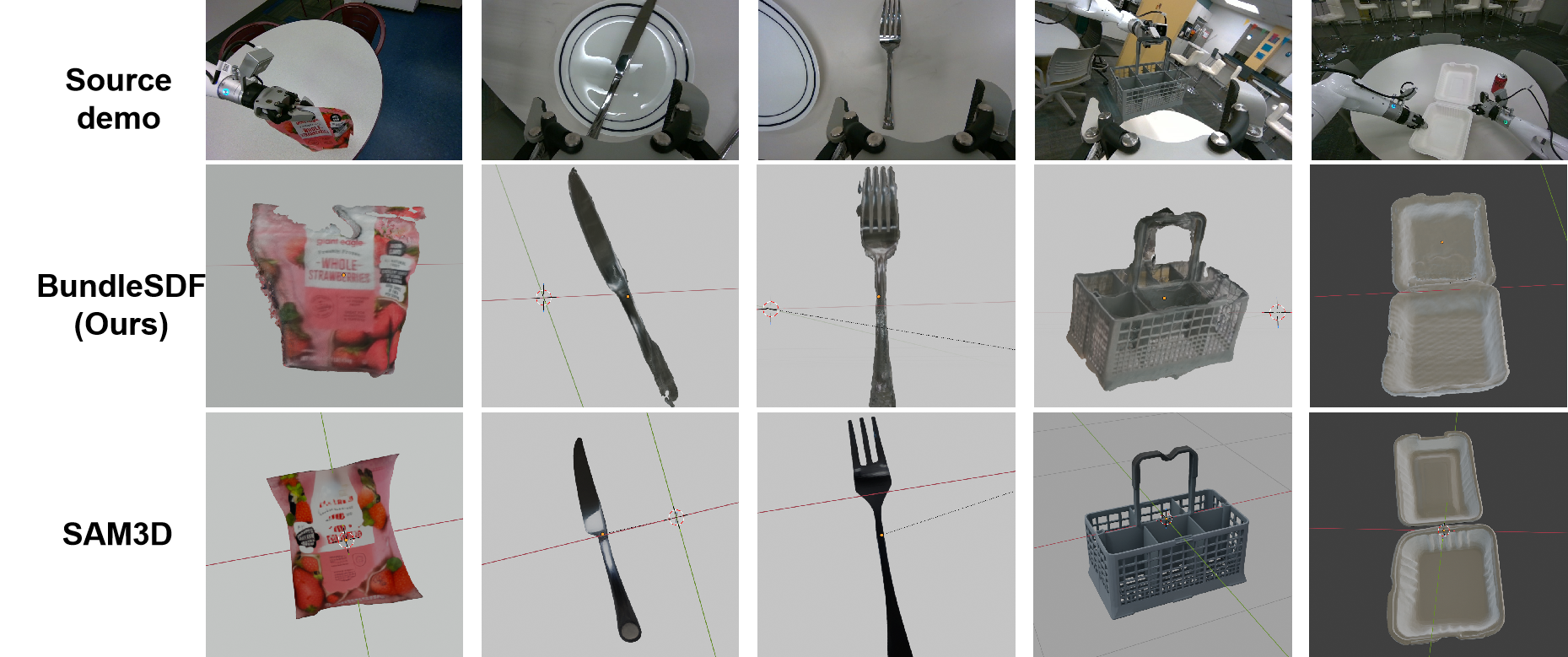}
  \caption{Additional foreground reconstruction results.}
  \label{fig:fg_recon_results}
\end{figure}

To validate the effectiveness of the generative completion method, we show comparisons of the object mesh generated by \textbf{BundleSDF}, \textbf{SAM3D} and \textbf{BundleSDF w/ Gen} method in Figure~\ref{fig:bundlesdf_gen_exp}. Note that we only apply this method to tasks that do not contain the complete observation over the object. As shown in Figure \ref{fig:bundlesdf_gen_exp}, our approach achieves both complete geometry shape and precise texture reconstruction. Due to incomplete observation, \textbf{BundleSDF} fails to get reasonable shape at the unseen faces. SAM3D fails to generate photo-realistic textures, while our approach can even recover the text on the objects, as shown in the first row of Figure \ref{fig:bundlesdf_gen_exp} and the first column of Figure \ref{fig:fg_recon_results}.

For demonstrations that contains complete observation over the object, we directly adopt \textbf{BundleSDF} and provide a comparison with \textbf{SAM3D} in Figure \ref{fig:fg_recon_results}. Compared with the image-to-3D methods that are commonly used in previous real2sim work~\citep{mao2025robot,jain2025polaris}, our method achieves photo-realistic texture reconstruction and precise geometry reconstruction with minimal sim-and-real gap. 

For robot, we utilize the robot URDF or USD to directly replay the joint state for contact-rich segments recorded during data collection. We utilize Isaac Sim to render both robots and objects, align the background light color and apply with minimal light brightness domain randomization to simulate the light changes in the real environment.
\label{app:recon_fg}
% TODO: foreground (object) reconstruction details, including the
% generative completion of unseen object faces.

% Figure 10 (wider via trimming) and Figure 11 (match Figure 10 width).

\begin{figure}[H]
  \centering
  \includegraphics[width=\linewidth,trim=3mm 0mm 3mm 0mm,clip]{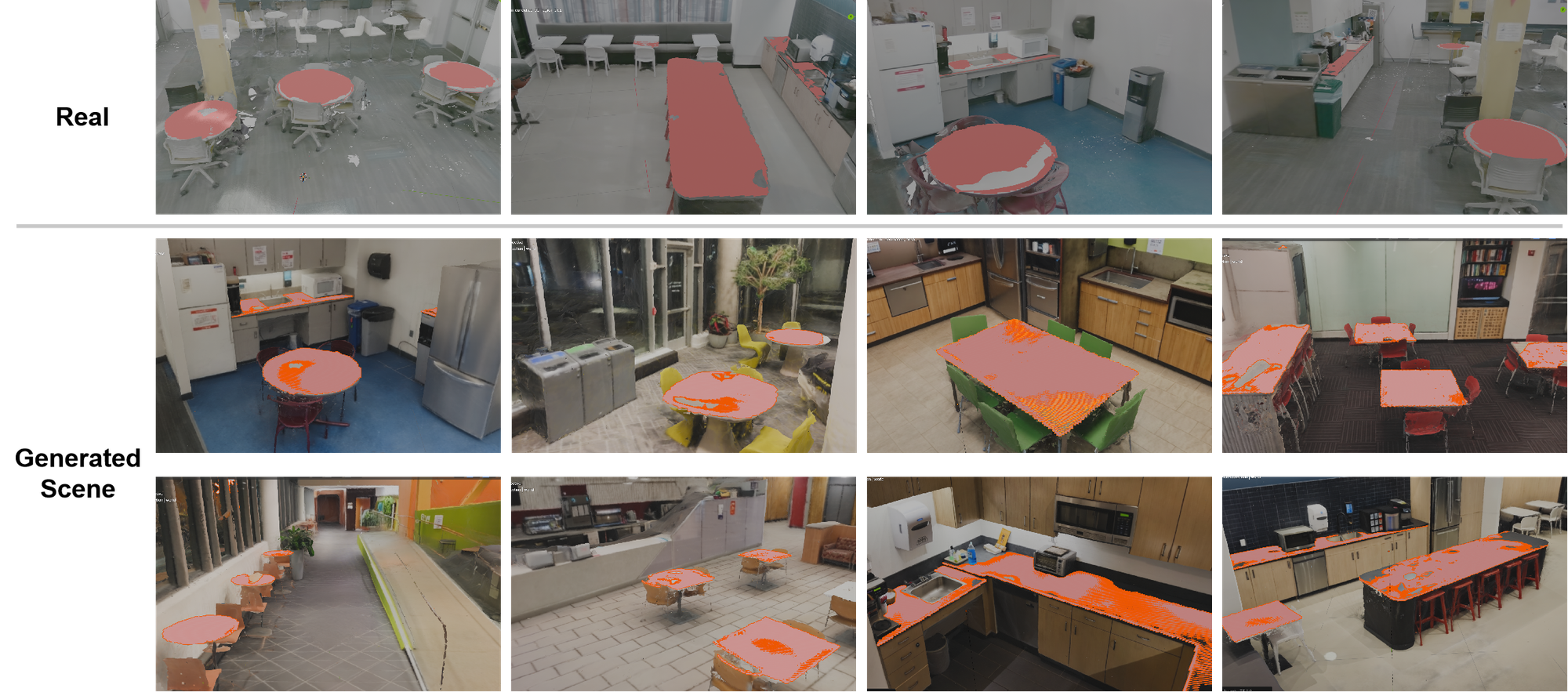}
  \caption{Initial configuration results on reconstructed and generated 3D scenes.}
  \label{fig:initial_config}
\end{figure}

\subsection{VLM-guided Initial Configuration}
\label{app:initial_config}
We provide the detailed prompt used in Appendix~\ref{initial_config_prompt}. Here we show the results of initial configuration on both reconstructed 3D scenes and generated 3D scenes. The VLM successfully recognize the reasonable region for randomizing the objects, and place the objects in extensive spaces to enhance the spatial coverage of the generated data. 

\section{Novel Trajectory Generation}
\label{app:trajgen}

\paragraph{Notation.} Homogeneous transforms $\mathbf{T}^{B}_{A}\in\mathrm{SE}(3)$ map frame $B$ into frame $A$; $W$ is the world frame and $o$ the target-object frame, with object position $\mathbf{p}_o$. The mobile base is the planar state $\mathbf{b}=[x_b,y_b,\psi_b]^\top\in\mathrm{SE}(2)$. End-effector (EE) targets are $\mathbf{T}^{E_k}_W$ for the left/right arm $k\in\{\mathrm{L},\mathrm{R}\}$, with rotation and translation blocks $\mathbf{R}^{E_k}_W,\mathbf{p}^{E_k}_W$ and constant object-relative grasp $\mathbf{T}^{E_k}_o=(\mathbf{T}^{o}_W)^{-1}\mathbf{T}^{E_k}_W$. $\mathbf{R}_z(\cdot)\in\mathrm{SO}(3)$ is a world-$z$ yaw rotation, $\hat{\mathbf{d}}$ a unit direction, $(\cdot)^\star$ an IK-achieved quantity, and the superscript $\mathrm{src}$ marks source-demonstration quantities. All weights and tolerances ($w_\bullet,\theta_{\max},c_{\min},\epsilon_p,\epsilon_R,\mathcal{R}_t,\mathcal{R}_\psi,\alpha,R$) are configuration constants, while $\Psi_{\mathrm{yaw}}$ below is a path quantity. The navigation endpoint is $p_e$ with yaw $\psi_e$ and unit approach direction $\hat{\mathbf{d}}_{\mathrm{app}}$; $\hat{\mathbf{d}}_{\mathrm{head}}$ is the head forward axis, $\hat{\mathbf{d}}_{H\to o}$ the head-to-object direction, $\mathbf{p}_{\mathrm{piv}}$ the centroid of the perturbed object group, and $\mathcal{K}\subseteq\{\mathrm{L},\mathrm{R}\}$ the set of tracked arms.

\subsection{Whole-Body Motion Planning}
\label{app:planning}
For each contact-rich segment, the planner regenerates a complete whole-body trajectory in the target world through three stages---navigation, arm approach, and replay---while preserving the segment itself verbatim in object-relative coordinates.

\paragraph{Navigation.} We first sample a set of base candidates on $\mathrm{SE}(2)$ around the target object (concentric rings and rays radiating from the source base pose). Each candidate is collision-checked (base footprint against the scene mesh of the world substrate) and has its visibility evaluated; visibility hard-rejects a candidate only when a strict visibility gate is enabled, and otherwise enters the ranking cost below. For surviving candidates we solve dual-arm whole-body IK to the segment's start EE poses with a multi-seed solver (seeded from the current and source postures), and plan a collision-free base path with RRT-Connect over $\mathrm{SE}(2)$, which is then retimed into a smooth turn--drive--turn motion. If line of sight is lost mid-route, the planner inserts a yaw-only spin to re-face the object and replans the remainder from the new heading, and for rotationally symmetric objects several object-yaw variants are tried.
A candidate is feasible ($\mathbf{b}\in\mathcal{B}_{\mathrm{feas}}$) when its retimed base path $P_{\mathbf{b}}$ passes the collision and IK gates (and the visibility gate, when configured). Each feasible path is scored by a weighted cost, and---by the configured policy---either the first feasible candidate or the lowest-cost one is committed:
\begin{equation}
\begin{aligned}
C_{\mathrm{vis}}(P)&=
w_\rho\bigl(1-\rho_{\mathrm{vis}}\bigr)
+\tfrac{w_c}{2}\bigl(1-\bar c\bigr)
+w_f\frac{|\theta_{\mathrm{err}}|}{\pi}
+w_{\mathrm{xy}}\,L_{\mathrm{xy}}
+w_\psi\frac{\Psi_{\mathrm{yaw}}}{\pi},\\
\mathbf{b}^\star&=
\begin{cases}
\text{first }\mathbf{b}\in\mathcal{B}_{\mathrm{feas}} & \text{(first-feasible policy)},\\[2pt]
\operatorname*{arg\,min}_{\mathbf{b}\in\mathcal{B}_{\mathrm{feas}}} C_{\mathrm{vis}}(P_{\mathbf{b}}) & \text{(min-cost policy)}.
\end{cases}
\end{aligned}
\label{eq:vis_cost}
\end{equation}
Over the $N$ retimed steps of the base path, $\hat{\mathbf{h}}_i$ is the head forward direction (the head camera's forward axis, or the planar base heading $[\cos\psi_i,\sin\psi_i]^\top$ as a fallback), the direction to the object is $\hat{\mathbf{d}}_i=\operatorname{normalize}(\mathbf{p}_o-\mathbf{p}_i)$, and the facing cosine is $c_i=\hat{\mathbf{h}}_i^\top\hat{\mathbf{d}}_i$; a step is visible when it clears the field-of-view, facing, and occlusion tests,
\begin{equation}
\mathrm{vis}_i=\mathbbm{1}\!\left[\arccos(c_i)\le\theta_{\max}\right]\,
\mathbbm{1}\!\left[c_i\ge c_{\min}\right]\,
\mathbbm{1}\!\left[\mathrm{ray}_i\ \text{clear}\right].
\label{eq:vis_step}
\end{equation}
Here $\rho_{\mathrm{vis}}=\tfrac1N\sum_i\mathrm{vis}_i$ is the visible ratio, $\bar c=\tfrac1N\sum_i c_i$ the mean facing cosine, $\theta_{\mathrm{err}}=\bigl|\operatorname{atan2}(p_{o,y}-y_b,\,p_{o,x}-x_b)-\psi_b\bigr|_{\mathrm{wrap}}$ the terminal facing error, $L_{\mathrm{xy}}=\sum_i\lVert\mathbf{p}_{i}^{xy}-\mathbf{p}_{i-1}^{xy}\rVert$ the travelled distance, and $\Psi_{\mathrm{yaw}}=\sum_i|\Delta\psi_i|_{\mathrm{wrap}}$ the cumulative yaw change. In contrast to a hard visibility constraint, this cost lets the planner trade off visibility against path effort when ranking feasible candidates.

\paragraph{Arm approach.} From the committed navigation pose, we plan a collision-free arm path with joint-space RRT-Connect to the segment's start EE pose.

\paragraph{Replay.} During the contact-rich segment, the per-frame dual-arm targets follow the standard object-relative (contact-relative) transfer of the source EE poses~\citep{mandlekar2023mimicgen,li2025momagengen}:
\begin{equation}
\mathbf{T}^{E_k}_{W}(i)=\mathbf{T}^{o}_{W}\bigl(\mathbf{T}^{o,\mathrm{src}}_{W}(i)\bigr)^{-1}\mathbf{T}^{E_k,\mathrm{src}}_{W}(i),
\qquad k\in\{\mathrm{L},\mathrm{R}\},
\label{eq:contact_transform}
\end{equation}
which we realize by per-frame whole-body IK while tracking the recorded object motion. Replaying in object-relative coordinates preserves the contact interaction exactly, while the surrounding trajectory is regenerated for the new spatial configuration.

\subsection{Corrective State Expansion}
\label{app:cse}
A planner-feasible trajectory follows a narrow, on-manifold distribution, whereas a real mobile robot deviates from it over long horizons: navigation overshoots, the arms drift while carrying, and the gripper approaches the target from unfamiliar angles. Corrective State Expansion (CSE) broadens state coverage at different stages, and every generated sample is re-validated for kinematic feasibility and collision before it is accepted.

\paragraph{Object State Expansion.} Applied inside the planner, this form simulates navigation drift relative to the object. With the base parked at $\mathbf{b}$ and frozen, the (grouped) target object is perturbed rigidly about the group centroid $\mathbf{p}_{\mathrm{piv}}$. A translation and yaw are drawn uniformly from configured (possibly asymmetric) ranges, applied in the world frame by default (optionally the base frame):
\begin{equation}
(\delta x,\delta y,\delta z)\sim\mathcal{U}(\mathcal{R}_t),\qquad
\delta\psi\sim\mathcal{U}(\mathcal{R}_\psi),
\label{eq:perturb_sample}
\end{equation}
\begin{equation}
\boldsymbol{\Delta}=
\begin{bmatrix}\mathbf{R}_z(\delta\psi) & \mathbf{p}_{\mathrm{piv}}+[\delta x,\delta y,\delta z]^\top-\mathbf{R}_z(\delta\psi)\,\mathbf{p}_{\mathrm{piv}}\\[2pt] \mathbf{0}^\top & 1\end{bmatrix}.
\label{eq:perturb_delta}
\end{equation}
The rigid transform $\boldsymbol{\Delta}\in\mathrm{SE}(3)$ rotates about $\mathbf{p}_{\mathrm{piv}}$; applying it to the object pose and propagating through the fixed grasp yields the updated targets
\begin{equation}
\tilde{\mathbf{T}}^{o}_{W}=\boldsymbol{\Delta}\,\mathbf{T}^{o}_{W},
\qquad
\mathbf{T}^{E_k}_{W}=\tilde{\mathbf{T}}^{o}_{W}\,\mathbf{T}^{E_k}_{o},\quad k\in\{\mathrm{L},\mathrm{R}\},
\label{eq:perturb_ee}
\end{equation}
which are re-solved by whole-body IK over an enlarged seed set (the perturbed target may leave the basin of the original seed). A sample is accepted only if every tracked arm ($k\in\mathcal{K}$) is within tolerance and the configuration is collision-free,
\begin{equation}
\mathbbm{1}_{\mathrm{feas}}=
\prod_{k\in\mathcal{K}}
\mathbbm{1}\!\left[\lVert\mathbf{p}^{E_k,\star}_{W}-\mathbf{p}^{E_k}_{W}\rVert\le\epsilon_p\right]\,
\mathbbm{1}\!\left[\arccos\!\Big(\tfrac{\operatorname{tr}(\mathbf{R}^{E_k,\star\top}_{W}\mathbf{R}^{E_k}_{W})-1}{2}\Big)\le\epsilon_R\right]\,
\mathbbm{1}\!\left[\text{collision-free}\right],
\label{eq:perturb_feas}
\end{equation}
optionally re-checking a few evenly spaced replay frames under the fixed base; among accepted candidates we keep the one minimizing the worst-arm position error. This widens the distribution of object positions seen during manipulation and improves spatial generalization.

\paragraph{Robot State Expansion.} This form perturbs the robot's own state at two stages. In the \emph{navigation} phase we apply two perturbations. First, a new start $p_s$ is sampled in an area-uniform cone opening backward from the approach direction $\hat{\mathbf{d}}_{\mathrm{app}}$ (cone axis $-\hat{\mathbf{d}}_{\mathrm{app}}$):
\begin{equation}
\begin{aligned}
p_s&=p_e+R\sqrt{u}\,\bigl[\cos\phi,\ \sin\phi\bigr]^\top,\qquad u\sim\mathcal{U}(0,1),\\
\phi&=\operatorname{atan2}\!\bigl(-\hat{d}_{\mathrm{app},y},\,-\hat{d}_{\mathrm{app},x}\bigr)+\Delta\theta,\qquad
\Delta\theta\sim\mathcal{U}(-\alpha,\alpha).
\end{aligned}
\label{eq:cone_sample}
\end{equation}
To preserve the object-relative endpoint pose, the object and endpoint are co-rotated about $\mathbf{p}_o$ by a closed-form yaw $\Delta\psi$ chosen so that, on arrival, the base heads from $p_s$ toward the rotated endpoint. With $D=\lVert p_s-p_o\rVert$, $r=\lVert p_e-p_o\rVert$, $\alpha_0=\operatorname{atan2}(p_{e,y}-p_{o,y},\,p_{e,x}-p_{o,x})$, $C=\psi_e-\alpha_0$, and $\varphi=\operatorname{atan2}(p_{o,y}-p_{s,y},\,p_{s,x}-p_{o,x})$, the orbit angle $\alpha^\star$ of the rotated endpoint satisfies
\begin{equation}
\sin(\alpha^\star+C+\varphi)=\frac{r\sin C}{D},
\qquad
\Delta\psi=(\alpha^\star-\alpha_0)_{\mathrm{wrap}},
\label{eq:corotate}
\end{equation}
selecting the forward-facing root; the object, endpoint, and both end-effectors are then advanced by $\mathbf{R}_z(\Delta\psi)$ about $\mathbf{p}_o$, and the robot drives a straight, retimed path from $p_s$. The start yaw is chosen so the target is visible at the start, i.e.\ $\mathbbm{1}[\arccos(\hat{\mathbf{d}}_{\mathrm{head}}^\top\hat{\mathbf{d}}_{H\to o})\le\theta_{\max}]\,\mathbbm{1}[\text{ray clear}]=1$ (with a fallback if none qualifies), and the trajectory is then accepted on its densified path's collision check alone:
\begin{equation}
\mathbbm{1}_{\mathrm{nav}}=\mathbbm{1}\!\left[\text{path collision-free}\right].
\label{eq:nav_feas}
\end{equation}
Second, at a random navigation frame we inject bounded Gaussian noise into the arm joints, graduated proximal-to-distal (small at the shoulder, larger at the wrist), followed by an ease-out recovery back to the carried pose, with any held object moved rigidly by the corresponding forward-kinematics delta; this teaches the policy to correct arm drift that accumulates while the robot is moving. In the \emph{manipulation} phase, the approach-to-object segment is replaced by a cone-cycloid EE trajectory~\citep{wang2025fieldgen} whose start is sampled inside a cone behind the target gripper pose and whose end matches the original target; per-frame inverse kinematics solves the active arm, and the trajectory is accepted only if every frame meets tight position and rotation tolerances and is collision-free. Together these expansions teach the policy not only to execute the nominal trajectory but to recover to it from the realistic drift that accumulates over long horizons.

\section{Progress Score and Subtask Evaluation}
\label{app:progress}
% TODO: definition of the progress score and full per-subtask success
% evaluation for each task.
\subsection{Progress Score Definition}
We provide the detailed definition of the progress score for each task here.
\begin{list}{\textbullet}{%
  \setlength{\leftmargin}{1.4em}% move list left
  \setlength{\labelsep}{0.5em}% space between bullet and text
}
  \item \textit{Lunch Box:} 
  This task is scored out of 4: \textbf{1.} the robot needs to successfully navigate to the lunch box and grasp the lunch box edge with left gripper; \textbf{2.} successfully close the lunch box using right gripper; \textbf{3.} open left gripper, retract left arm and succeed to grasp the closed lunch box with right gripper; \textbf{4.} take the lunch box, navigate to the kitchen sink and place it in the sink. 
  \item \textit{Utensil:} The task is scored out of 4: \textbf{1.} the robot searches for the knife and navigates to it; \textbf{2.} grasps the knife precisely with its right hand; \textbf{3.} carries the knife and navigates to the storage basket; \textbf{4.} places the knife in the basket.
  \item \textit{Drop Trash:} The task is scored out of 2. \textbf{1.} the robot needs to first search and navigate to the empty cola can; \textbf{2.} take the can, find the trash can, navigate to the trash can and throw it in.
  \item \textit{Pour:} The task is scored out of 2. \textbf{1.} it needs to first navigate to the teapot and grasp it; \textbf{2.} then take the teapot to the cup and pour into it.
  \item \textit{Fridge:} The task is scored out of 3. \textbf{1.} the robot needs to first navigate to the grocery bag and grasp it; \textbf{2.} navigate to the fridge, and store the bag in it; \textbf{3.} close the fridge door.
\end{list}

\subsection{Subtask Success Rate}
The per-subtask success rate for each real-world task is reported in Figure~\ref{fig:subtask_sr}, comparing \method with the ablation without Corrective State Expansion. As tasks run without resets, for subtask $k$ we report the fraction of rollouts that complete it among those that reach it (i.e.\ that completed subtask $k{-}1$), annotating each bar with the raw count; a method that never reaches a subtask has an undefined rate ($0/0$), drawn as a short stub on the axis. \method completes each individual subtask with high reliability, including the contact-rich late stages, whereas without Corrective State Expansion the policy is markedly weaker at every stage and often fails to reach the later ones. This shows Corrective State Expansion curbs the error accumulation of long-horizon mobile manipulation.

\subsection{Scene Generalization Subtask Success Rate}
Figure~\ref{fig:scenegen_subtask} breaks the scene-generalization evaluation in Figure~\ref{fig:scaling}(b) down by subtask. We train \method with Marble-generated worlds and evaluate it on four tasks in three real scenes ($10$ rollouts per task and scene), reporting the same per-subtask success rate as above. \method completes subtasks across all three novel scenes, confirming that synthesizing data on generated worlds transfers to the corresponding real environments.

\section{Multi-View Videos of Synthesized Data}
\label{app:videos}
\begin{figure*}[p]
  \centering
  \makebox[\textwidth][r]{\includegraphics[width=1.1116\textwidth]{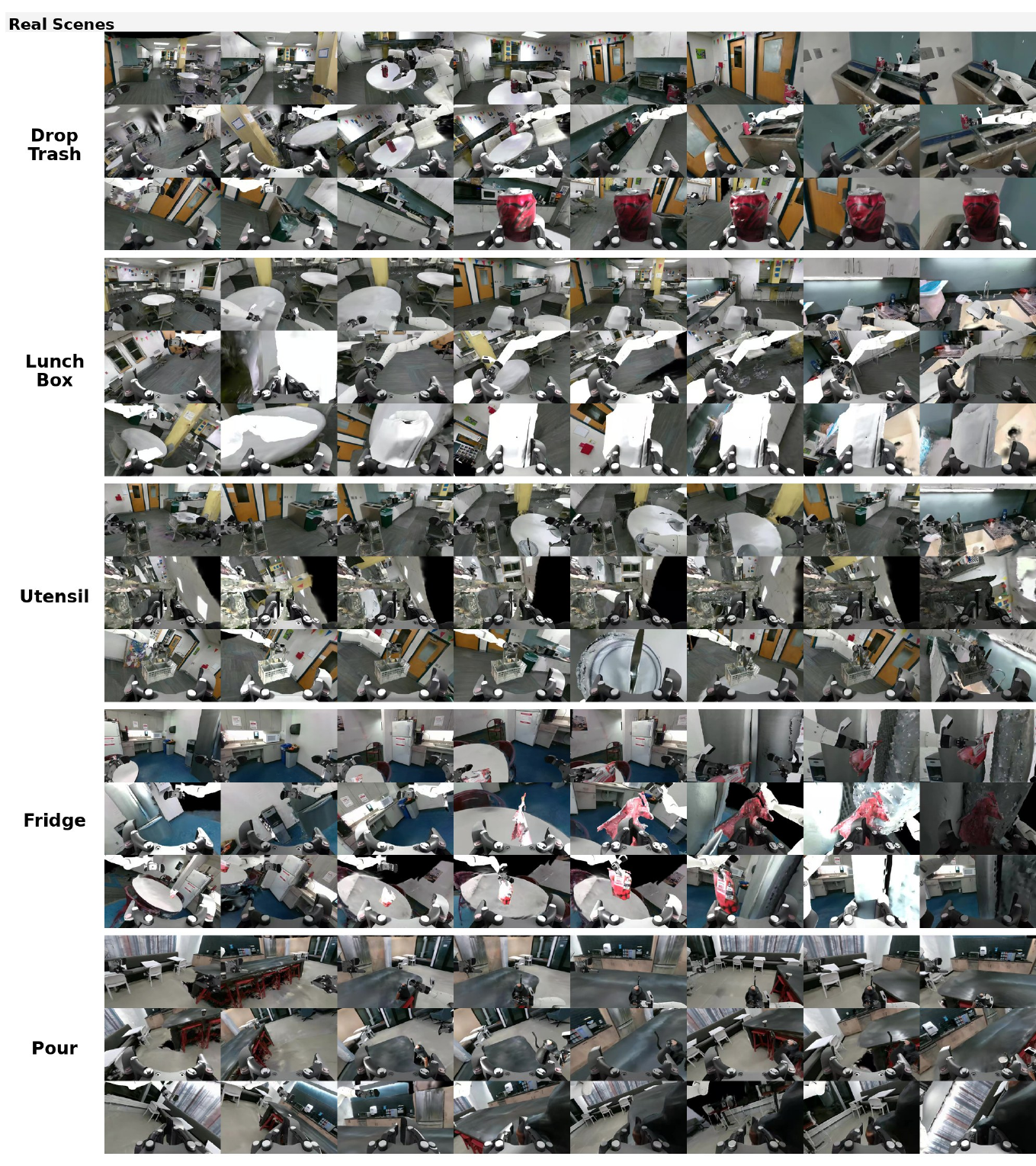}}
  \caption{Representative frames from the real-scene trajectories.}
  \label{fig:appendix-multiview-1}
\end{figure*}

\begin{figure*}[p]
  \centering
  \makebox[\textwidth][r]{\includegraphics[width=1.1116\textwidth]{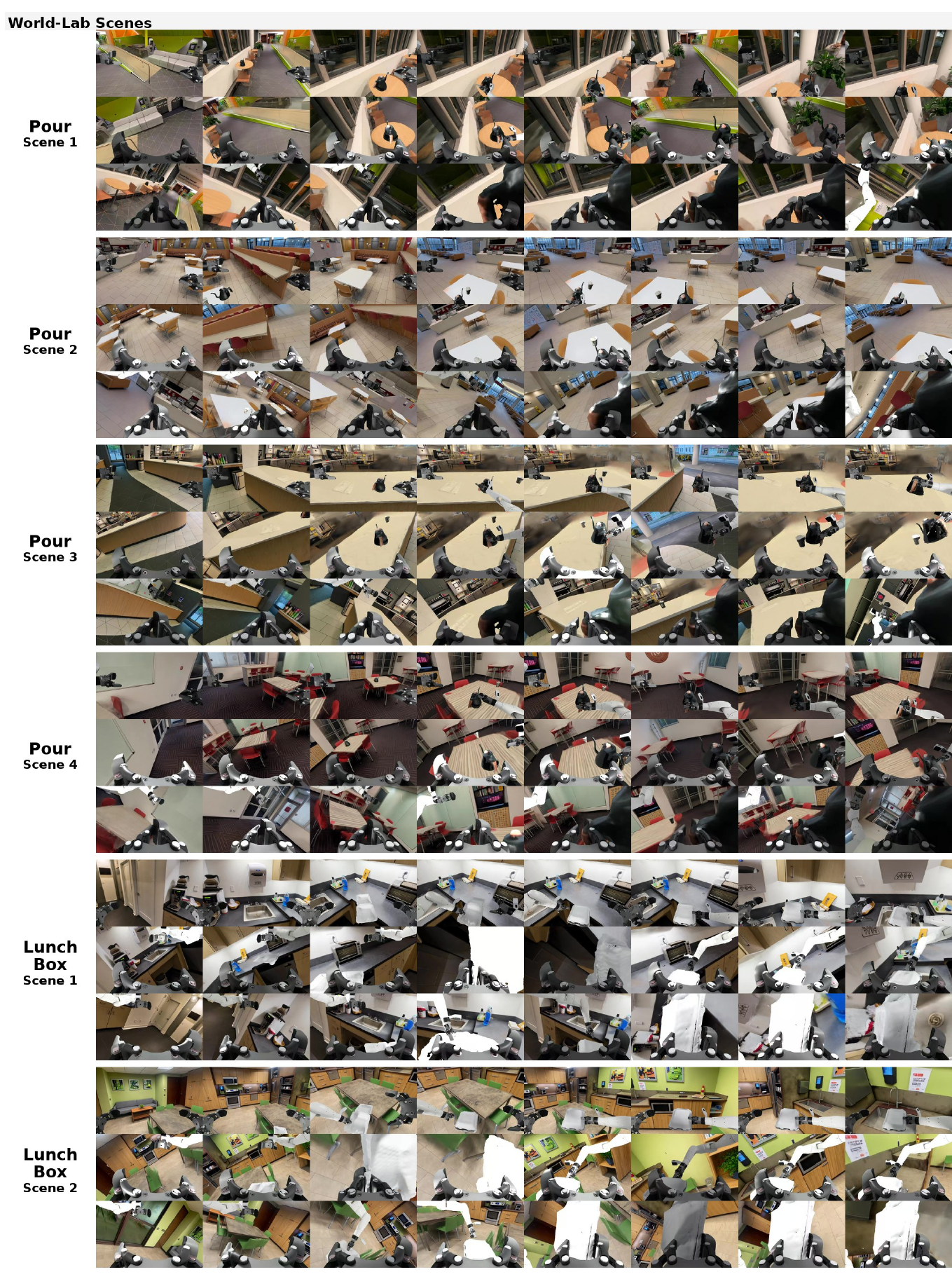}}
  \caption{Representative frames from the world-lab trajectories.}
  \label{fig:appendix-multiview-2}
\end{figure*}

\begin{figure*}[p]
  \centering
  \makebox[\textwidth][r]{\includegraphics[width=1.1116\textwidth]{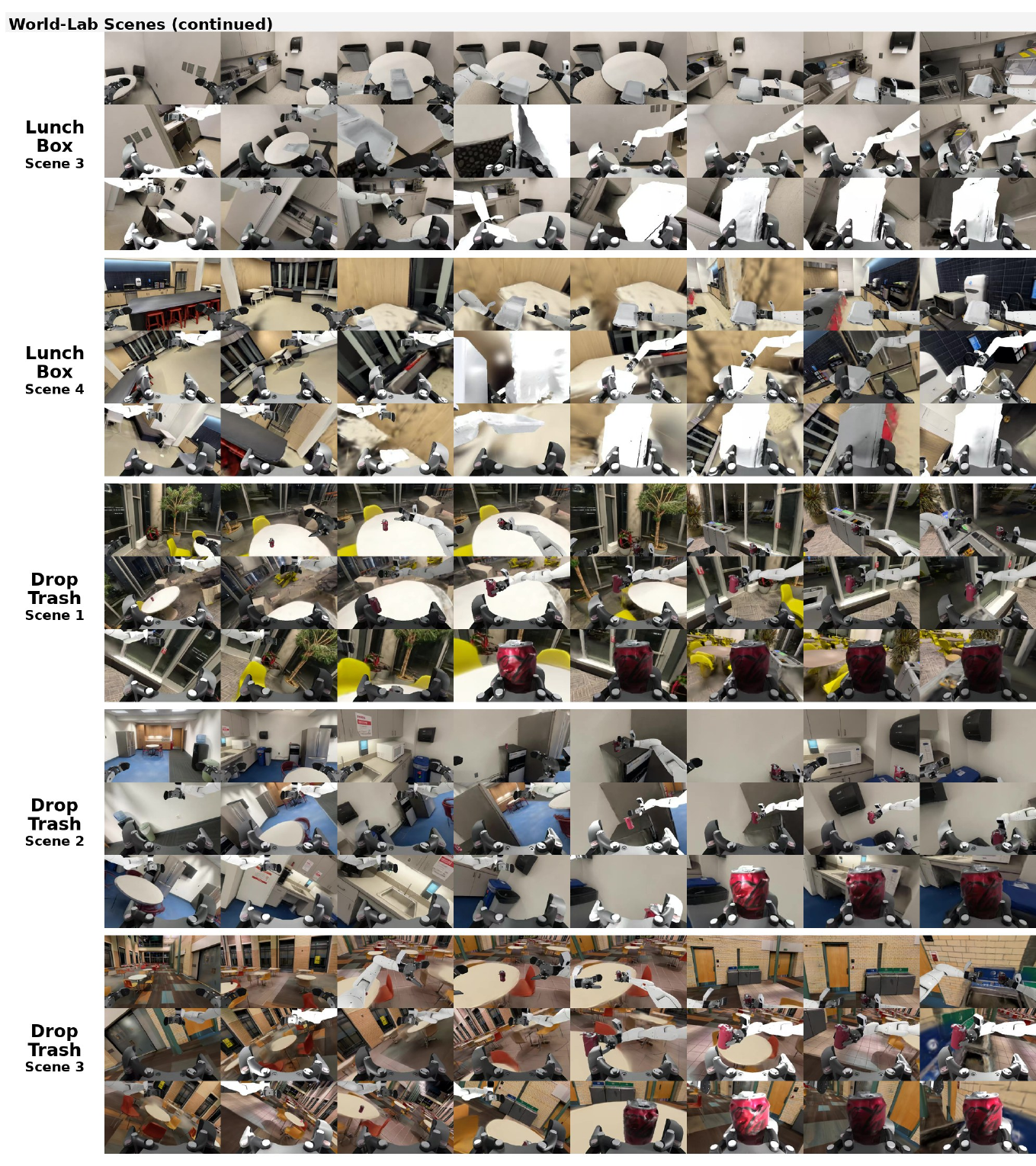}}
  \caption{Representative frames from the world-lab trajectories (continued).}
  \label{fig:appendix-multiview-3}
\end{figure*}

We show the visualization of head and wrist camera videos from our synthesized data in Figures~\ref{fig:appendix-multiview-1}--\ref{fig:appendix-multiview-3}. Our approach achieves realistic rendering of the visual observations at high resolution (head camera: $1280\times800$, wrist cameras: $800\times 480$). The gap between rendered and real observations is negligible at the lower $224\times 224$ resolution fed into the policy. For each trajectory we show eight frames sampled along the trajectory for the head and the two wrist cameras. Figure~\ref{fig:appendix-multiview-1} shows the reconstructed real scenes for all five tasks, and Figures~\ref{fig:appendix-multiview-2} and~\ref{fig:appendix-multiview-3} show the world-lab scenes for \textit{Pour}, \textit{Lunch Box}, and \textit{Drop Trash}.

% Frames span the full text width; the row labels overhang into the (empty) left margin.

% --- Least important; remove this section if the per-task cost is dropped ---
\section{Per-Task Data-Generation Cost}
\label{app:cost}
Table~\ref{tab:app_cost} breaks down the compute to generate each real-world dataset, for the reconstructed source scene and the Marble-generated worlds. \emph{Data} is the duration of the generated 30\,fps multi-view video; \emph{Plan} is wall-clock planning time on a single 192-core AMD EPYC 9R14 node; \emph{FG} and \emph{BG} are the foreground (Isaac~Sim) and background (Gaussian-splat) render time, and \emph{Render} their sum, in L40S GPU-hours. In total, generating $251.6$ data-hours costs $24.6$ planning hours and $765$ GPU-hours.

\begin{table}[h]
\centering
\small
\caption{\textbf{Per-task data-generation cost.}}
\label{tab:app_cost}
% Data-generation cost table snippet (meant to be \input{} inside a table*)
\setlength{\tabcolsep}{5pt}
\begin{tabular}{llrrrrrr}
\toprule
Task & Scene & \shortstack{Data\\(h)} & \shortstack{Plan\\(h)} & \shortstack{FG\\(GPU-h)} & \shortstack{BG\\(GPU-h)} & \shortstack{Render\\(GPU-h)} & \shortstack{GPU-h\\/\,data-h} \\
\midrule
\taskname{Lunch Box}  & Source  & 15.3 & 0.2 & 28  & 34 & 62  & 4.05 \\
                      & Worlds  & 85.0 & 4.3 & 153 & 69 & 222 & 2.61 \\
\taskname{Drop Trash} & Source  & 13.0 & 0.4 & 23  & 26 & 50  & 3.81 \\
                      & Worlds  & 24.4 & 9.7 & 44  & 20 & 64  & 2.61 \\
\taskname{Pour}       & Source  & 17.8 & 0.6 & 32  & 36 & 68  & 3.84 \\
                      & Worlds  & 49.7 & 4.2 & 89  & 40 & 130 & 2.61 \\
\taskname{Utensil}    & Source  & 37.0 & 3.4 & 67  & 68 & 134 & 3.63 \\
\taskname{Fridge}     & Source  & 9.4  & 1.7 & 17  & 18 & 35  & 3.72 \\
\midrule
\textbf{Total} &     & \textbf{251.6} & \textbf{24.6} & \textbf{453} & \textbf{311} & \textbf{765} & \textbf{3.04} \\
\bottomrule
\end{tabular}

\end{table}

Foreground rendering is robot-only and thus task-independent. Background rendering is cheaper for generated worlds, whose lighter splats render about $3\times$ faster, lowering their cost to $2.61$ GPU-hours per data-hour against $3.6$--$4.1$ for source scenes. Planning accounts for only a small fraction of the total cost.

\section{Cross-embodiment Data Generation}
\label{app:cross_emb}
\begin{figure}[H]
  \centering
  % Slightly overhang into the margin (similar trick as Fig.\,16) to make it larger.
  \makebox[\textwidth][r]{\includegraphics[width=1.14\textwidth]{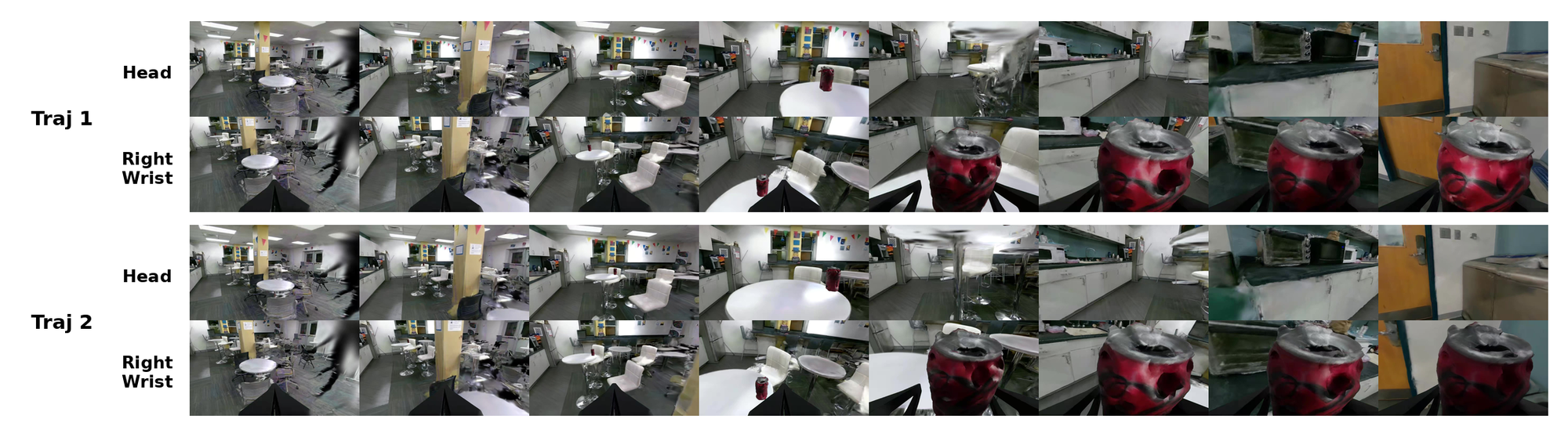}}
  \caption{Cross-embodiment data generation visual observations from Agibot to Linearbot.}
  \label{fig:cross_emb_data}
\end{figure}

Figure~\ref{fig:cross_emb_data} shows the generated visual observations for the cross-embodiment data generation from Agibot~G1 to Linearbot on \textit{Drop Trash}, described in Section~\ref{sec:cross_emb}. Despite a large kinematic gap (different DoFs) and a large visual gap (different camera extrinsics and intrinsics) between the two robots, \method generates reasonable visual observations.

\section{Policy Training Details}
\label{app:training}
We train our policies by full-parameter fine-tuning of $\pi_{0.5}$~\citep{physicalintelligence2025pi05} from the public \texttt{pi05\_base} checkpoint on data generated by \method. $\pi_{0.5}$ is a PaliGemma-based vision--language--action model---a Gemma-2B vision--language backbone with a Gemma-300M action expert. The policy consumes three $224\times224$ RGB views (the head camera and the two wrist cameras), the robot proprioceptive state, and a tokenized language instruction (up to $200$ tokens), and predicts an action chunk of horizon $32$. The proprioceptive-state and action dimensionality is $20$ for the dual-arm Agibot~G1, padded to the model's $32$-dimensional action space.

When fine-tuning $\pi_{0.5}$, we use AdamW ($\beta_1{=}0.9$, $\beta_2{=}0.95$, $\epsilon{=}10^{-8}$, weight decay $10^{-10}$, global gradient-norm clip $1.0$) with a cosine-decay learning-rate schedule (peak $5\times10^{-5}$, $1{,}000$-step warmup, decaying to $2.5\times10^{-6}$). We train for $30{,}000$ steps at batch size $256$ on $4$ nodes of $8\times$NVIDIA H100 GPUs, taking roughly $12$ hours. Training uses mixed precision: the trainable weights and AdamW state are kept in \texttt{float32}, while the forward and backward passes compute in \texttt{bfloat16}. Table~\ref{tab:train_hparams} summarizes the full set of hyperparameters.

\begin{table}[h]
\centering
\small
\caption{\textbf{$\pi_{0.5}$ fine-tuning hyperparameters.}}
\label{tab:train_hparams}
\begin{tabular}{ll}
\toprule
Hyperparameter & Value \\
\midrule
Initialization        & \texttt{pi05\_base} checkpoint \\
Fine-tuning           & full-parameter \\
Training steps        & 30{,}000 \\
Batch size            & 256 \\
Optimizer             & AdamW ($\beta_1{=}0.9$, $\beta_2{=}0.95$, $\epsilon{=}10^{-8}$, wd $10^{-10}$) \\
Gradient-norm clip    & 1.0 \\
LR schedule           & cosine decay \\
Peak / final LR       & $5\times10^{-5}$ / $2.5\times10^{-6}$ \\
Warmup steps          & 1{,}000 \\
Action chunk horizon  & 32 \\
Precision             & mixed (\texttt{bf16} compute, \texttt{fp32} weights) \\
\bottomrule
\end{tabular}
\end{table}

Each task is specified by a fixed natural-language instruction that conditions the policy throughout the episode. Table~\ref{tab:task_prompts} lists the instructions for the five real-world tasks.

\begin{table}[h]
\centering
\small
\caption{\textbf{Language instructions for the real-world tasks.}}
\label{tab:task_prompts}
\begin{tabular}{l p{0.72\linewidth}}
\toprule
Task & Language instruction \\
\midrule
\taskname{Lunch Box}  & ``Close the lunch box on the table, pick it up, and throw it into the sink.'' \\
\taskname{Utensil}    & ``Pick up the knife from the plate and place it in the basket.'' \\
\taskname{Drop Trash} & ``Pick up the cola can on the table and drop it into the recycling bin.'' \\
\taskname{Fridge}     & ``Pick up the bag on the table, put it in the refrigerator, and close the refrigerator.'' \\
\taskname{Pour}       & ``Pick up the teapot and pour water into the cups.'' \\
\bottomrule
\end{tabular}
\end{table}

\section{Prompt}
\subsection{Inpaint Prompt}
\label{inpaint_prompt}
\paragraph{Inpainting Prompt.}
The inpainting model is instructed to remove all bright green masked regions and any visible robot parts, including arms, grippers, torso, mobile base, links, or other robot body components. It must reconstruct the missing background so that it appears natural, seamless, and consistent with the surrounding pixels. The prompt explicitly requires that pixels outside the removed masked or robot regions remain unchanged.

\paragraph{Judging Prompt.}
A separate reviewer model evaluates the generated result by comparing it with the original image. The reviewer checks whether:
\begin{enumerate}
    \item all bright green masked regions were fully removed;
    \item all visible robot parts were fully removed;
    \item the filled background looks natural and seamless;
    \item no unintended changes occurred outside the original masked or robot regions.
\end{enumerate}
The reviewer must output either \texttt{KEEP} or \texttt{RETRY}, followed by a concise reason.

\paragraph{Corrective Prompt.}
If the result fails review, a corrective prompt is generated using the previous inpainting prompt, the original image, the failed result, and the failure reason. The corrected prompt focuses on fixing the observed failure while preserving the core constraints: removing green masks and robot parts, completing the background naturally, and avoiding edits outside the masked or robot regions.

\subsection{Initial Configuration Prompt}
\label{initial_config_prompt}
\paragraph{Task Randomization Prompt.}
The pipeline takes a user-provided natural-language task description, such as
\texttt{pick up the cola can}, and uses it as the main task prompt for selecting
scene regions relevant to randomized manipulation. This task description is
passed to the task-driven reasoning stage, which identifies candidate objects,
surfaces, or spatial regions that are important for the requested task.

\paragraph{Task-Driven Region Selection.}
Given rendered RGB-D views of the scene and the task description, the pipeline
runs task-driven reasoning, segmentation, backprojection, and top-$k$ region
selection. The goal is to find the most task-relevant randomization regions in
the 3D scene. The selected regions are saved in a manifest file containing the
top-ranked candidate regions.

\paragraph{Scene Rendering Context.}
Before task reasoning, the pipeline renders multiple RGB-D views of the input
mesh. For WorldLab scenes, orbit views are rendered using configurable camera
parameters such as azimuth count, field of view, orbit origin, and up-axis. For
real scenes, camera poses are loaded from a corresponding \texttt{cameras.json}
file.

\paragraph{Scale and Packaging Prompt Context.}
For non-real scenes, the pipeline estimates scene scale using a vision-language
model. The support surface is described by the configurable label
\texttt{PIPELINE\_\allowbreak SUPPORT\_\allowbreak LABEL}, whose default value is
\texttt{tabletop surface}. This label guides the scale-estimation stage when
interpreting the selected scene regions.

\paragraph{Final Output.}
The pipeline packages the rendered images, depth maps, selected and scaled
regions, visualization files, and debug candidates into the final output
directory. The retained outputs include:
\begin{itemize}
    \item \texttt{images/}, \texttt{depth/}, and \texttt{depth\_mm/};
    \item \texttt{orbit\_views.json};
    \item \texttt{scaled\_assets/} and \texttt{scaled\_regions/};
    \item \texttt{visualization/} and \texttt{debug\_candidates/}.
\end{itemize}

\end{document}